\documentclass[10pt,twocolumn,letterpaper]{article}

\usepackage[pagenumbers]{wacv} 

\definecolor{wacvblue}{rgb}{0.21,0.49,0.74}
\usepackage[pagebackref,breaklinks,colorlinks,allcolors=wacvblue]{hyperref}

\def\wacvPaperID{*****} 
\def\confName{WACV}
\def\confYear{2027}

\title{FAVE: Foveated Adaptive Visual Encoding for Efficient Fine-Grained Visual Understanding}

\author{
Amitangshu Mukherjee \qquad Kaushik Roy\\
Elmore Family School of Electrical and Computer Engineering\\
Purdue University\\
{\tt\small mukher44@purdue.edu \qquad kaushik@purdue.edu}
}

\begin{document}
\maketitle

\begin{abstract}
Fine-grained visual understanding depends on local detail, yet visual encoders face a trade-off between costly full-image high-resolution processing and compact global encoding that can weaken such evidence. Inspired by human active vision, we separate \emph{where} to look from \emph{what} to encode. We focus on the latter and introduce \textbf{FAVE} (\textbf{F}oveated \textbf{A}daptive \textbf{V}isual \textbf{E}ncoding), a lightweight variable-resolution ViT that encodes externally selected regions at high acuity while preserving native geometry. We first isolate this encoding problem using oracle ground-truth crops in a controlled small-object regime. On ImageNet objects with a native maximum side of 96 pixels, FAVE improves Top-1 by 9.4 points over a fixed-resolution ViT on the same crop window with \(12.7\times\) lower FLOPs. Increasing global resolution or backbone capacity does not recover the same operating point. We then integrate FAVE as a complementary local branch in FastVLM. Its local tokens are combined with FastVLM's global visual tokens, while the original global pathway and language model remain frozen. With at most 16 additional local tokens, FAVE improves TextVQA by 1.60 points and achieves a \(3.3\times\) controlled TTFT speedup over SmolVLM2-2.2B. On GQA attribute questions, it improves FastVLM-1.5B by 1.31 points, extending the benefit beyond text while narrowing the gap to FastVLM-7B. Together, these results show that selectively allocating high-acuity local capacity provides an efficient complement to broader global representations and model scaling for fine-grained understanding of small objects, text, and attributes.
\end{abstract}

\section{Introduction}
\label{sec:intro}

Image classification has long served as a fundamental task for visual recognition~\cite{deng2009imagenet,russakovsky2015imagenet}. As visual perception has expanded to object detection, segmentation, structured scene understanding, and now vision--language reasoning, preserving task-relevant visual evidence has become increasingly demanding~\cite{Lin2014MicrosoftCC,krishna2017visualgenome}. In modern vision--language systems, visual encoders provide the perceptual representation that language models use for multimodal reasoning~\cite{alayrac2022flamingo,li2023blip2,liu2023llava}. This evidence is inherently multi-scale. Global scene structure provides context, while fine local details distinguish small objects, attributes, text, and other task-relevant evidence. A useful visual system must preserve both, yet doing so does not require uniformly high spatial resolution across the image.

\begin{figure*}[t]
    \centering
    \includegraphics[
        width=\textwidth,
        trim=0 1cm 0 0,
        clip
    ]{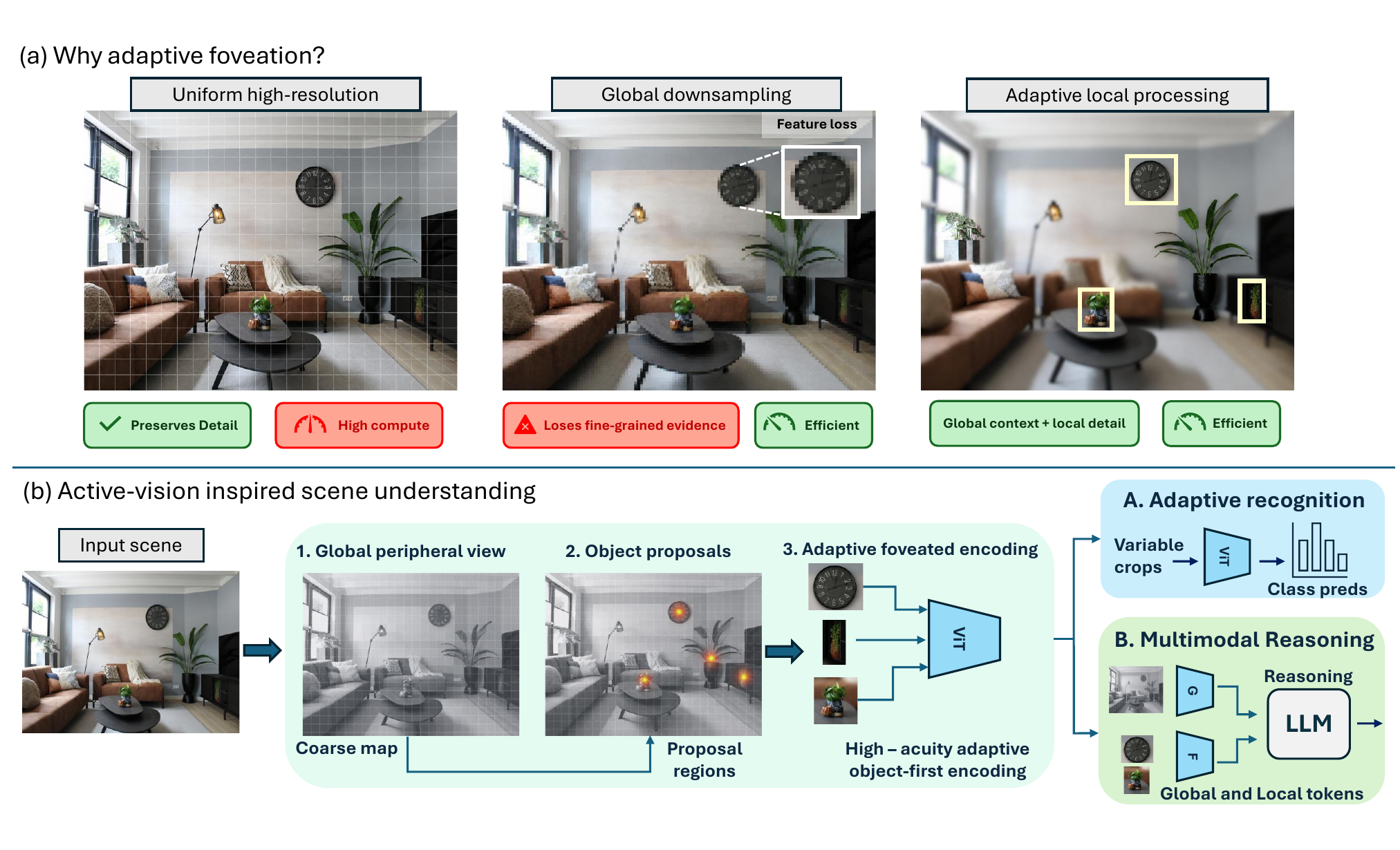}
    \caption{
    \textbf{Motivation and overview of adaptive foveated visual processing.}
    (a) Uniform high-resolution processing preserves fine-grained visual evidence but incurs high computational cost, whereas global downsampling is efficient but can discard locally important details. Selective high-resolution processing retains global context while allocating additional computation to salient local regions.
    (b) Our approach builds on active vision: a coarse peripheral representation preserves global scene context, candidate regions provide potential gaze targets, and selected regions receive adaptive high-acuity processing. The resulting global and foveated local representations support both adaptive visual recognition and multimodal reasoning.\protect\footnotemark
    }
    \label{fig:overview}
    \vspace{-4mm}
\end{figure*}

A straightforward way to preserve fine detail is to process the complete image at high resolution. For Vision Transformers (ViTs), however, higher resolution rapidly increases the visual-token count and the cost of global self-attention~\cite{dosovitskiy2021an,SwinV2,vaswani2017attention}. With a fixed patch size, doubling both image dimensions produces four times as many visual tokens, including regions where such detail may be unnecessary. Aggressive global downsampling offers the opposite trade-off: it reduces computation but can discard small yet task-relevant evidence before downstream processing. Figure~\ref{fig:overview}(a) summarizes this tension between visual detail and computational efficiency. Selective high-resolution processing provides a middle ground, retaining global context while directing additional visual computation only to regions that require finer detail. Selective processing also allows us to add a lightweight foveated encoder for these regions, rather than enlarging the entire full-image visual backbone.

\footnotetext{Source photograph: Mendy Van Hoogdalem van Barneveld, Pexels. \url{https://www.pexels.com/photo/photo-of-a-living-room-13675290/}}

This selective allocation of visual computation is closely related to active vision. Human perception does not sample an entire scene uniformly. Peripheral vision provides coarse global context, while task-driven saccadic eye movements direct the high-acuity fovea toward regions that require finer inspection~\cite{clark2013whatever,goodale2004sight,milner1992visual,mishkin1983object}. This motivates a distinction between \emph{where} to look and \emph{what} to encode: spatial selection identifies regions for closer inspection, while high-acuity foveated processing preserves the visual evidence within them~\cite{ungerleider1994what,eckstein2011visual}. Figure~\ref{fig:overview}(b) translates this principle into a computational setup: a global pathway preserves scene context, an external selector determines where to inspect, and selected regions receive high-acuity local processing. As illustrated by the couch and clock, these regions can differ substantially in scale and therefore need not receive the same visual budget. We make the local representation adaptive, allowing spatial resolution and token count to vary across selected regions while keeping computation bounded.

This leads to \textbf{FAVE}, our foveated adaptive visual encoder. Once a region has been selected, FAVE preserves its native geometry rather than forcing it to a common square resolution. Small regions therefore remain computationally small, while larger regions can use more visual computation within the budget. A variable-resolution vision transformer with factorized positional embeddings supports these region-dependent inputs, while Patch-n-Pack efficiently batches the resulting variable-length sequences~\cite{navit}. These mechanisms enable FAVE, but the central idea is adaptive high-acuity encoding that preserves fine-grained evidence within selected visual regions.

We first study FAVE in isolation through object recognition. ImageNet~\cite{deng2009imagenet} provides a controlled setting for asking how a selected object should be encoded once its location is known. We use ground-truth bounding boxes as oracle spatial selections, separating \emph{where} to look from \emph{how} to encode what is selected. This isolates the accuracy--compute trade-off of adaptive local encoding and lets us test where additional visual computation is most effective. The same need for fine local evidence arises in vision--language models. Broad scene context may be sufficient for many questions, while others depend on small objects, text, or attributes that can be weakened by compact global encoding. A language model can reason over the evidence it receives, but it cannot reliably reconstruct visual information that was not preserved by the visual encoder. We use FastVLM~\cite{fastvlm} as the global pathway because it provides a compact visual-token sequence for efficient scene understanding. FAVE complements this representation by encoding selected regions at higher acuity. The resulting local representations are compressed into a small token budget and added to the FastVLM global tokens. The FastVLM global pathway and language model remain frozen, isolating the contribution of the local visual pathway. Together, these settings examine one encoder-side question: can fine-grained evidence within selected regions be preserved through targeted local visual capacity while retaining an efficient global representation? ImageNet isolates how to encode a selected region, while the VLM setting tests how that encoding can complement a compact global pathway with limited token and latency overhead.
The main contributions of this work are as follows:
\begin{itemize}

\item We introduce \textbf{FAVE}, a foveated adaptive visual encoder for selected regions. FAVE preserves native geometry and adapts spatial resolution and token count to region scale, avoiding a fixed-resolution representation for every region.

\item We isolate this encoding principle on ImageNet using oracle spatial selection. On objects with native maximum side $\leq 96$ pixels, FAVE improves Top-1 by 9.4 points over a fixed-resolution ViT on the same crop window with 12.7$\times$ lower FLOPs. The 5.7M-parameter FAVE-T also outperforms substantially larger fixed-resolution ViTs in this regime.

\item We integrate FAVE as a sparse local complement to FastVLM's efficient global representation. FastVLM retains global scene context, while FAVE contributes a small set of local visual tokens. The original global pathway and language model remain frozen.

\item On TextVQA, FAVE improves the FastVLM-1.5B baseline by 1.60 points with at most 16 additional local tokens. With FAVE augmentation, FastVLM-1.5B becomes competitive with compact VLMs such as SmolVLM2-2.2B, while reducing controlled TTFT by a factor of 3.3 relative to SmolVLM2-2.2B.

\item We extend the evaluation beyond text to fine-grained object attributes on GQA. FAVE improves FastVLM-1.5B by 1.31 points with only a 0.42\% increase in total parameters, reaching 72.04\% attribute accuracy, within 0.68 points of FastVLM-7B.
\end{itemize}

\section{Related Work}
\label{sec:rwork}

We review related work in three areas: foveated vision, adaptive visual encoding, and efficient multimodal perception. Broader discussion is provided in Appendix~A.

\subsection{Foveated vision}

Computational foveation models peripheral vision through eccentricity-dependent blur, pooling, and non-uniform sampling~\cite{akbas2014object,deza2016peripheral,deza2021emergent,shah2023training,Poggiocortical}. FocL instead processes object-centric crops, but at a fixed resolution~\cite{mukherjee2026from}, while Segment This Thing uses a fixed point-centered foveation pattern with progressively coarser peripheral patches~\cite{schmidt2025segment}. Given externally selected regions, FAVE preserves their native geometry, allowing local resolution to adapt to each region without imposing a fixed-resolution representation.

\begin{figure*}[t]
    \centering
    \includegraphics[width=0.95\textwidth]{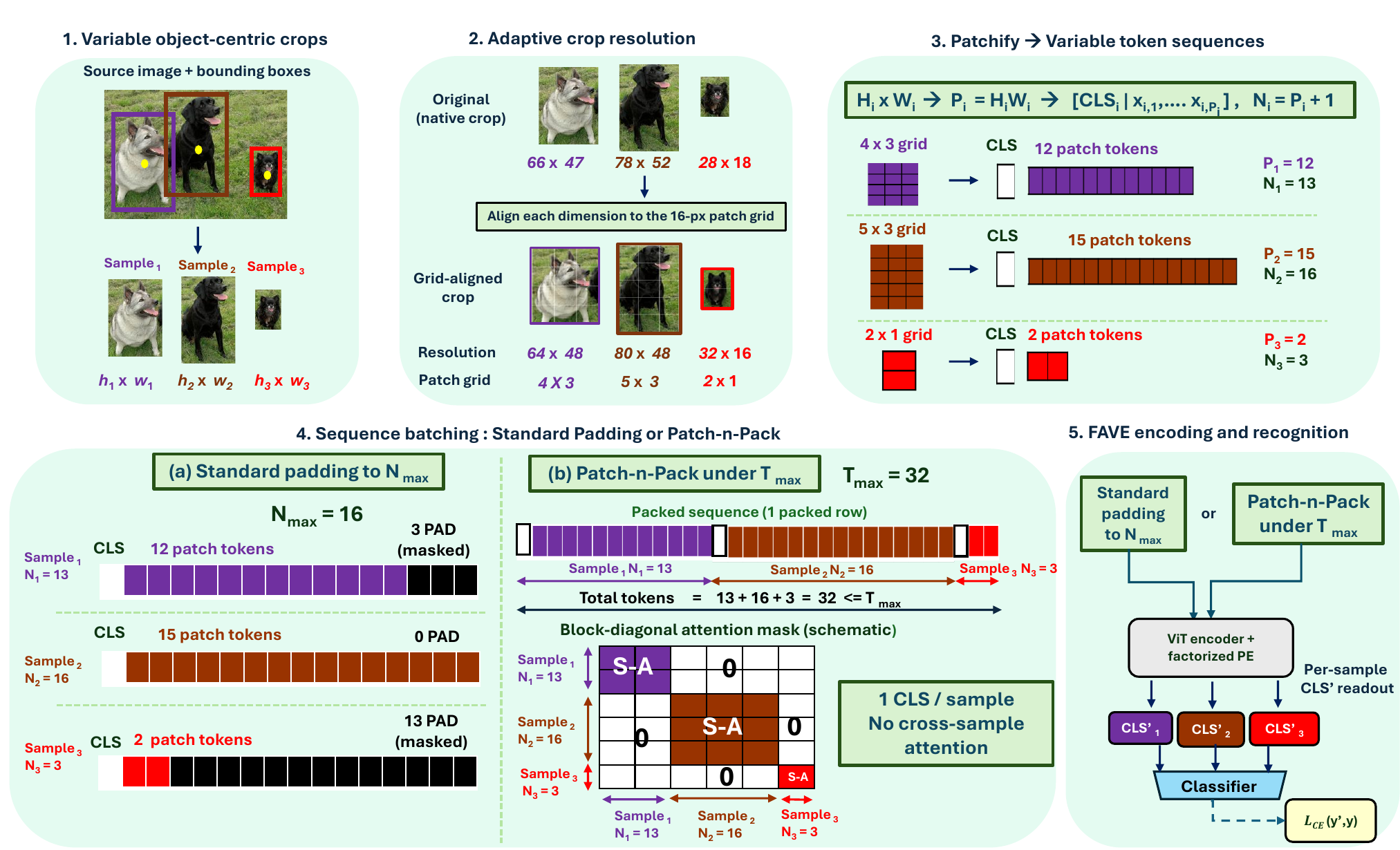}
    \caption{
    \textbf{FAVE: adaptive object-centric visual encoding for recognition.}
Starting from selected object-centric crops, FAVE preserves aspect ratio and
aligns each crop to the 16-pixel patch grid, producing variable-resolution
patch grids. Each crop yields \(P_i = H_iW_i\) patch tokens and sequence length
\(N_i = P_i + 1\) after prepending a CLS token. Variable-length sequences are
processed either with standard padding to \(N_{\max}\) or with
Patch-n-Pack under token budget \(T_{\max}\), using block-diagonal
self-attention to prevent cross-sample interaction. A shared ViT with
factorized positional encoding produces one CLS representation per crop for
recognition.
    }
    \label{fig:fave_imagenet}
\end{figure*}

\subsection{Efficient and Adaptive Visual Encoding}

Efficient visual encoders reduce cost through architectural design, variable-resolution processing, or token reduction. Swin Transformer, Focal Transformer, and FocalNet structure attention across spatial scales~\cite{SwinV2,yang2021focal,yang2022focal}, while EfficientViT and FastViT target efficient inference~\cite{liu2023efficientvit,vasufastvit2023}. ResFormer and ViTAR support variable or high-resolution inputs~\cite{tian2022resformer,fan2024vitar}, whereas DynamicViT, Adaptive Token Sampling, and Token Merging prune, sample, or merge visual tokens~\cite{rao2021dynamicvit,Fayyaz2021AdaptiveTS,bolya2022tome}. NaViT preserves native image resolution and aspect ratio, with Patch-n-Pack batching variable-length sequences~\cite{navit}. FAVE applies variable-resolution processing to externally selected regions rather than the full image, preserving native geometry and adapting the local spatial budget without token pruning or learned selection.

\subsection{Efficient and Adaptive Multimodal Perception}

Compact VLMs reduce multimodal cost through smaller language backbones, efficient visual encoders, or compact visual-token sequences. SmolVLM2 and TinyLLaVA target small-scale multimodal models, while FastVLM uses FastViTHD to reduce vision latency and produce a compact global visual representation~\cite{marafioti2025smolvlm,tinyllava,fastvlm}. TokenPacker and VisionZip further reduce the visual sequence before language-model inference~\cite{li2025tokenpacker,yang2025visionzip}.

A complementary line of work targets fine-grained perception by allocating additional visual detail when the global representation is insufficient. LLaVA-UHD uses adaptive image slicing, while AttWarp, ViCrop, SD-RPN, and Q-Zoom refine or revisit informative image regions~\cite{guo2024llavauhd,dalal2025constructive,zhang2025mllms,shi2026sdrpn,shi2026qzoom}. Visual CoT and Foveated Reasoner incorporate localized visual evidence into the reasoning process, while LLMind adapts image sampling under a constrained pixel budget~\cite{shao2024visual,min2026foveatedreasoning,debnath2026llmind}.

Multi-encoder VLMs instead combine complementary visual representations. Mini-Gemini adds a second visual encoder for high-resolution refinement without increasing the visual-token count, while Cambrian-1 studies combinations of multiple vision encoders and spatially aggregates their high-resolution features~\cite{MiniGemini,tong2024cambrian}. Our VLM instantiation instead uses complementary visual pathways with a different division of labor: FastVLM retains the efficient global representation, while a lightweight FAVE encoder re-encodes externally selected regions. Spatial selection remains separate from local encoding, and the resulting local representations are compressed into a small token budget while the global pathway and language model remain frozen.

\section{Foveated Adaptive Visual Encoding}
\label{sec:fave}

We first define FAVE as an adaptive encoder for selected visual regions, then
use ImageNet with ground-truth spatial selection fixed to isolate the
encoder-side question: once a region is known, how should it be encoded?

\begin{table*}[t]
\centering
\small
\setlength{\tabcolsep}{3.0pt}
\begin{tabular}{@{}llccc ccc ccc@{}}
\toprule
& &
\multicolumn{3}{c}{Native bbox $\leq 96$} &
\multicolumn{3}{c}{Native bbox $\leq 224$} &
\multicolumn{3}{c}{General-2K} \\
\cmidrule(lr){3-5}
\cmidrule(lr){6-8}
\cmidrule(l){9-11}
Model & Input
& T1 $\uparrow$ & Tok. $\downarrow$ & G $\downarrow$
& T1 $\uparrow$ & Tok. $\downarrow$ & G $\downarrow$
& T1 $\uparrow$ & Tok. $\downarrow$ & G $\downarrow$ \\
\midrule

Fixed-Res ViT-T/16
& Image
& 32.65 & 197.0 & 1.283
& 46.30 & 197.0 & 1.283
& 61.75 & 197.0 & 1.283 \\

& Oracle crop
& 41.05 & 197.0 & 1.283
& 55.70 & 197.0 & 1.283
& 64.60 & 197.0 & 1.283 \\

\addlinespace

Adaptive ViT-T (Full)
& Image
& 23.80 & 149.3 & 0.943
& 29.25 & 153.3 & 0.969
& 43.25 & 156.4 & 0.991 \\

& Oracle crop
& 9.55 & 17.7 & 0.101
& 29.35 & 71.6 & \textbf{0.434}
& 42.20 & 127.3 & \textbf{0.800} \\

\addlinespace

NaViT-T
& Image
& 34.25 & 148.3 & 0.958
& 43.45 & 152.3 & 0.985
& 59.55 & 155.4 & 1.007 \\

& Oracle crop
& 31.60 & \textbf{16.7} & \textbf{0.098}
& 49.60 & \textbf{70.6} & 0.439
& 60.35 & \textbf{126.3} & 0.812 \\

\addlinespace

\textbf{FAVE}
& Oracle crop
& \textbf{50.45} & 17.7 & 0.101
& \textbf{60.55} & 71.6 & \textbf{0.434}
& \textbf{65.60} & 127.3 & \textbf{0.800} \\

\bottomrule
\end{tabular}

\caption{
\textbf{Adaptive object recognition under fixed spatial selection.}
We compare full-image processing with GT-selected object regions across native
object-scale cohorts. Ground-truth boxes fix spatial selection so that the
object-centric rows isolate how the selected region is encoded. FAVE preserves
crop geometry and adapts its spatial resolution and token count, whereas
Fixed-Res ViT-T/16 resizes every selected crop to \(224\times224\).
T1, Tok., and G denote Top-1 accuracy, mean visual tokens per image, and
analytical GFLOPs per image. Bold indicates the best value within each cohort.
}
\label{tab:imagenet_main}
\end{table*}
\begin{table}[h]
\centering
\small
\setlength{\tabcolsep}{3.5pt}
\begin{tabular}{@{}llccc@{}}
\toprule
Model & Policy & T1 $\uparrow$ & Tok. $\downarrow$ & G $\downarrow$ \\
\midrule
\multicolumn{5}{c}{Native bbox $\leq 96$} \\
\midrule
FAVE & Native
& 50.45 & \textbf{17.7} & \textbf{0.101} \\
FAVE & Adaptive zoom
& \textbf{54.00} & 40.0 & 0.234 \\
NaViT-T & LS cap
& 34.25 & 148.3 & 0.958 \\
NaViT-T & \(r=224\)
& 34.20 & 192.3 & 1.277 \\
\midrule
\multicolumn{5}{c}{Native bbox $\leq 224$} \\
\midrule
FAVE & Native
& 60.55 & \textbf{71.6} & \textbf{0.434} \\
FAVE & Adaptive zoom
& \textbf{62.20} & 97.6 & 0.603 \\
NaViT-T & LS cap
& 43.45 & 152.3 & 0.985 \\
NaViT-T & \(r=224\)
& 43.25 & 192.3 & 1.277 \\
\bottomrule
\end{tabular}
\caption{
\textbf{Selective local magnification versus additional global resolution.}
FAVE allocates additional resolution to the GT-selected region using
scale-dependent magnification, whereas NaViT increases resolution over the
full image. T1, Tok., and G denote Top-1 accuracy, mean visual tokens per
image, and analytical GFLOPs per image.
}
\vspace{-2mm}
\label{tab:imagenet_allocation}
\end{table}

\vspace{-4mm}
\paragraph{Object-centric adaptive crops.} For the ImageNet instantiation in Fig.~\ref{fig:fave_imagenet}, FAVE operates
on object regions defined by dataset bounding boxes. Each selected region is
cropped directly from the source image and retains its native spatial dimensions
and aspect ratio rather than being resized to a common resolution. Small crops
remain at native resolution, while larger crops may optionally be isotropically
downscaled under a task-specific spatial or compute cap. The resulting crop is
then aligned to the \(16\)-pixel patch grid by adjusting its spatial boundaries
rather than resizing it, preserving crop-dependent geometry and producing
variable-size patch grids.

\vspace{-2mm}

\paragraph{Patch tokenization and positional encoding.} Each grid-aligned crop is partitioned into \(16\times16\) patches. Because crops
are not normalized to a common spatial size, differences in scale and aspect
ratio produce variable patch grids of size \(H_i\times W_i\), with
\(P_i = H_iW_i\) patch tokens. We prepend one \(\mathrm{CLS}_i\) token to form
\[
[\mathrm{CLS}_i \mid \mathbf{x}_{i,1},\ldots,\mathbf{x}_{i,P_i}],
\qquad N_i = P_i + 1.
\]
Since \(H_i\) and \(W_i\) vary across samples, the encoder must support
different grid shapes and aspect ratios. Following NaViT~\cite{navit}, we use
factorized positional encoding. For a patch at row \(u\) and column \(v\),
\[
\mathbf{e}^{\mathrm{pos}}_{u,v}
=
\mathbf{e}^{\mathrm{row}}_{u}
+
\mathbf{e}^{\mathrm{col}}_{v}.
\]
This avoids fixed-grid positional embeddings, allowing one ViT to process variable patch grids while region scale determines token count and compute.

\paragraph{Sequence batching.} Figure~\ref{fig:fave_imagenet}, Part~4, illustrates batching strategies for
FAVE's variable-length sequences. With standard batching, sequences are padded
to the longest length \(N_{\max}\); padding positions are masked, excluded from
valid self-attention interactions, and do not contribute to the task objective.
Alternatively, Patch-n-Pack~\cite{navit} packs complete sequences under a token
budget \(T_{\max}\), using block-diagonal attention to prevent cross-sample
interaction. Both preserve the per-sample token sequence, allowing the same
FAVE checkpoint to be executed with either batching scheme. Further details are
provided in Appendix B.

\vspace{-3mm}

\paragraph{FAVE encoding and recognition.} As shown in Fig.~\ref{fig:fave_imagenet}, Part~5, all variable-length crop
sequences are processed by a shared ViT encoder. The output
\(\mathrm{CLS}'_i\) representation is passed to a shared classifier and trained
with standard cross-entropy loss. The main ImageNet comparison uses a Tiny-scale ViT; larger fixed-resolution backbones are evaluated separately.

\begin{table}[t]
\centering
\small

\begin{tabular*}{\columnwidth}{@{\extracolsep{\fill}}lcrrr@{}}
\toprule
Model & Input & Params & T1 $\uparrow$ & GFLOPs $\downarrow$ \\
\midrule

ViT-T/16
& Full
& 5.72M
& 32.65
& 1.283 \\

ViT-T/16
& GT
& 5.72M
& 41.05
& 1.283 \\

\midrule

ViT-S/16
& Full
& 22.05M
& 43.50
& 4.657 \\

ViT-S/16
& GT
& 22.05M
& 45.65
& 4.657 \\

\midrule

ViT-B/16
& Full
& 86.57M
& 45.55
& 17.680 \\

ViT-B/16
& GT
& 86.57M
& 47.35
& 17.680 \\

\midrule

\textbf{FAVE-T (ours)}
& \textbf{GT}
& \textbf{5.70M}
& \textbf{50.45}
& \textbf{0.101} \\

\bottomrule
\end{tabular*}

\caption{
\textbf{Does increasing backbone capacity recover the small-object operating point?}
Fixed-resolution ViTs are evaluated on the full image (\emph{Full}) and
GT-selected crop (\emph{GT}) on the native-bbox $\leq96$ cohort.
FAVE-T uses adaptive native-geometry encoding of the same selected region.
}
\vspace{-4mm}
\label{tab:imagenet_capacity_scaling}
\end{table}

\begin{figure*}[t]
    \centering
    \includegraphics[
        width=0.95\textwidth,
        trim=0 2.5cm 0 0.4cm,
        clip
    ]{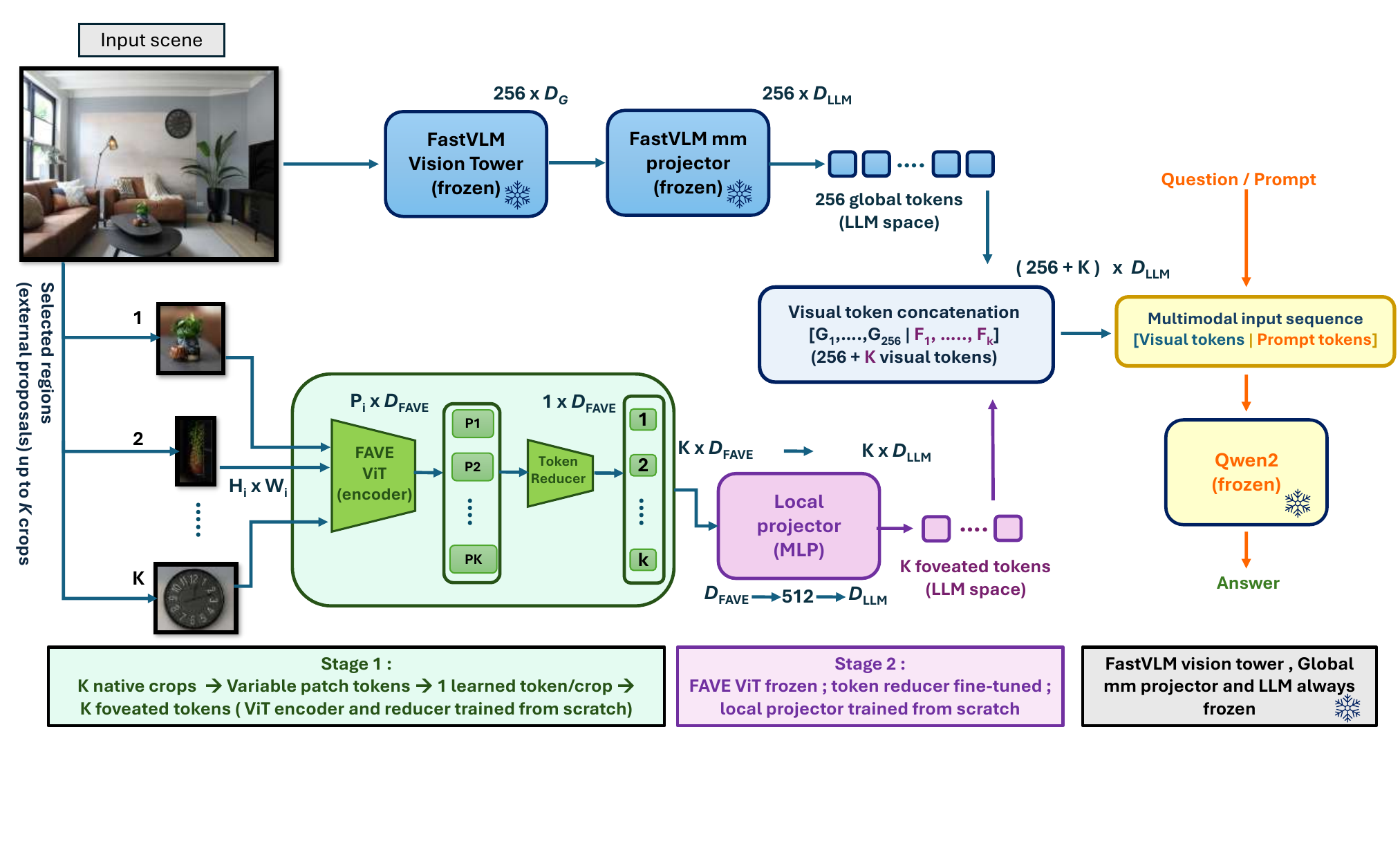}
    \caption{
    \textbf{FAVE for multimodal visual understanding.}
    The full image is processed by the frozen FastVLM vision tower and projector to produce 256 global visual tokens, while up to $K$ externally selected regions are encoded by FAVE. In Stage~1, the FAVE ViT and learned reducer are trained jointly from scratch to compress each native-geometry crop from $P_i$ variable-length features to one $D_{\mathrm{FAVE}}$-dimensional token. In Stage~2, the ViT is frozen, the reducer is fine-tuned, and a learned local projector maps the $K$ crop tokens into the $D_{\mathrm{LLM}}$ space. These local tokens are concatenated with the 256 FastVLM tokens and, together with the prompt, processed by the frozen Qwen2 language model. $D_G$, $D_{\mathrm{FAVE}}$, and $D_{\mathrm{LLM}}$ denote the global feature, FAVE feature, and language-model embedding widths, respectively.
    }
    \vspace{-4mm}
    \label{fig:fave_fastvlm}
\end{figure*}

\subsection{Adaptive Object Recognition}
\label{sec:imagenet}

\paragraph{Dataset and training.}
We use ImageNet images with bounding-box annotations and a fixed 90/10
train--validation split shared across methods. The main comparison
uses Tiny-scale ViT backbones with 12 blocks, embedding dimension \(D=192\),
and \(16\times16\) patches; larger fixed-resolution backbones are evaluated
separately in Table~\ref{tab:imagenet_capacity_scaling}. FAVE is trained on object-centric crops, whereas the full-image controls are trained on the corresponding complete images.
Optimization and dataset-construction details are provided in Appendices ~B and~C.

\vspace{-4mm}
\paragraph{Evaluation design.}
We use ImageNet as a recognition diagnostic for how a selected region should
be encoded once its location is known. Ground-truth boxes fix spatial selection,
removing proposal quality as a confound and enabling comparison on the same
region. We evaluate three independently sampled 2,000-image cohorts: native
bounding-box maximum side \(\leq96\) pixels, \(\leq224\) pixels, and an
unrestricted General-2K cohort. Table~\ref{tab:imagenet_main} separates the
effects of spatial selection, variable-resolution encoding, and object-centric
training. \textbf{Fixed-Res ViT-T/16} compares full-image processing with the
same GT-selected crop resized to \(224\times224\), testing the selected region
under a fixed visual budget. \textbf{Adaptive ViT-T (Full)} uses the same
variable-resolution backbone and, under GT-selected evaluation, the same crop
geometry and compute as FAVE, but is trained on complete images.
\textbf{NaViT-T} provides a generic variable-resolution control~\cite{navit}.
Together, these baselines isolate how the selected region is encoded.
Table~\ref{tab:imagenet_allocation} then asks where additional visual compute
should be allocated, while Table~\ref{tab:imagenet_capacity_scaling} asks
whether larger fixed-resolution ViTs recover the same accuracy--efficiency
operating point. We report Top-1 accuracy, mean visual tokens, and analytical
GFLOPs; further details are provided in Appendix~C.

\vspace{-5mm}

\paragraph{Adaptive local encoding.}
Table~\ref{tab:imagenet_main} shows that GT-selected regions help even under
fixed-resolution encoding: Fixed-Res ViT-T/16 improves from \(32.65\) to
\(41.05\) Top-1 on the \(\leq96\) cohort at unchanged compute. FAVE reaches
\(50.45\), a further \(9.4\)-point gain with \(11.1\times\) fewer tokens and
\(12.7\times\) lower analytical FLOPs. This is not explained by region access
alone: Adaptive ViT-T (Full) receives the same crop geometry and inference
budget but performs substantially worse, isolating the role of object-centric
training. NaViT-T also fails to recover FAVE accuracy, showing that generic
variable-resolution processing is insufficient. FAVE remains strongest on the
broader \(\leq224\) and General-2K cohorts, reaching \(60.55\) and \(65.60\)
Top-1.

\vspace{-5mm}

\paragraph{Selective compute allocation.}
Table~\ref{tab:imagenet_allocation} asks where additional visual compute should
be allocated. Scale-dependent magnification improves FAVE by \(3.55\) Top-1
points on the \(\leq96\) cohort and \(1.65\) points on the \(\leq224\)
cohort. In contrast, increasing NaViT's full-image resolution raises token
count and FLOPs without improving Top-1 on either cohort. Thus, at these
operating points, additional compute is more effective when targeted toward the
selected region rather than spread across the image.

\vspace{-5mm}

\paragraph{Backbone capacity.}
Table~\ref{tab:imagenet_capacity_scaling} asks whether larger fixed-resolution
backbones can recover the same small-object operating point. On the
\(\leq96\) cohort, scaling the GT-crop ViT from Tiny to Base raises Top-1
from \(41.05\) to \(47.35\), still below FAVE-T at \(50.45\).
FAVE-T uses only \(5.70\)M parameters and \(0.101\) GFLOPs, compared with
\(86.57\)M and \(17.68\) GFLOPs for ViT-B/16. Together with
Table~\ref{tab:imagenet_allocation}, these results show that targeted local
capacity provides a more efficient route to fine-grained evidence than broadly
increasing resolution or backbone capacity at these operating points.

Together, this section establishes FAVE as an efficient adaptive encoder for fine-grained evidence in selected regions. We next use it as a complementary local pathway alongside FastVLM's compact global representation.

\section{FAVE for Multimodal Visual Understanding}
\label{sec:glvlm}

We next test whether FAVE can complement FastVLM's compact global representation with sparse local evidence.

\noindent\textbf{Global and foveated visual pathways.}
The ImageNet results show that targeted local visual capacity can recover fine-grained evidence efficiently. We therefore ask whether the same principle can complement an efficient global VLM representation. As shown in Fig.~\ref{fig:fave_fastvlm}, FastVLM provides compact scene-level context through its frozen vision tower and multimodal projector, producing 256 global visual tokens~\cite{fastvlm}. In parallel, a task-specific external proposal mechanism selects up to $K$ source-image regions, which FAVE independently re-encodes to preserve fine-grained local evidence. Spatial selection determines where to look; FAVE determines how the selected regions are encoded.

\paragraph{Stage 1: Compact local token learning.}
Each selected crop is independently encoded by FAVE using the variable-resolution formulation of Sec.~3, producing $P_i$ patch features
$\mathbf{X}_i \in \mathbb{R}^{P_i \times D_{\mathrm{FAVE}}}$,
where $P_i$ varies with crop geometry. A shared learned cross-attention reducer compresses these features into one content-dependent crop token,
\[
\mathbf{z}_i = \mathcal{R}(\mathbf{X}_i),
\qquad
\mathbf{z}_i \in \mathbb{R}^{D_{\mathrm{FAVE}}}.
\]
Stage~1 trains the FAVE ViT and reducer jointly from scratch. Each region may use a different spatial resolution and number of patch features, but still contributes only one compact token. This decouples local encoding resolution from the downstream language-model token budget. The architecture is shared across multimodal settings; task-specific training objectives are provided in Appendix D.

\paragraph{Stage 2: Local projection and token fusion.}
Stage~2 maps the compact crop tokens into the language-model embedding space using a local projector $\mathcal{P}$,
\[
\mathbf{F}
=
\mathcal{P}(\mathbf{Z}),
\qquad
\mathbf{F}
\in
\mathbb{R}^{K \times D_{\mathrm{LLM}}}.
\]
The projected local tokens are then concatenated with the 256 global FastVLM tokens,
\[
\mathbf{V}
=
[\mathbf{G};\mathbf{F}]
\in
\mathbb{R}^{(256+K)\times D_{\mathrm{LLM}}},
\]
and combined with the task prompt for answer generation by the frozen Qwen2 language model. During Stage~2, the FAVE ViT is frozen, the token reducer is fine-tuned, and the local projector is trained from scratch. The FastVLM vision tower, global projector, and language model remain frozen.

\begin{table}[t]
\centering
\small

\begin{tabular*}{\columnwidth}{@{\extracolsep{\fill}}lr@{}}
\toprule
Method & TextVQA $\uparrow$ \\
\midrule

\multicolumn{2}{c}{Compact VLMs} \\
\midrule
SmolVLM-2.2B~\cite{marafioti2025smolvlm}
& 72.1$^{\dagger}$ \\
SmolVLM2-2.2B~\cite{marafioti2025smolvlm}
& 73.21$^{\dagger}$ \\
MobileVLM V2-1.7B~\cite{mobilevlmv2}
& 52.1$^{\dagger}$ \\
MobileVLM V2-3B
& 57.5$^{\dagger}$ \\
DeepSeek-VL-1.3B~\cite{deepseekvl}
& 55.8$^{\dagger}$ \\
Florence-VL~\cite{florencevl}
& 69.1$^{\dagger}$ \\

\midrule
\multicolumn{2}{c}{Multi-encoder VLMs (8B LLM)} \\
\midrule
MiniGemini-HD~\cite{MiniGemini}
& 70.2$^{\dagger}$ \\
Cambrian-1~\cite{tong2024cambrian}
& 71.7$^{\dagger}$ \\

\midrule
\multicolumn{2}{c}{Fine-grained perception: LLaVA-1.5-7B} \\
\midrule
LLaVA-1.5-~\cite{liu2023llava}
& 46.1$^{\ddagger}$ \\
\quad + S$^2$~\cite{shi2024when}
& 52.6$^{\ddagger}$ \\
\quad + ViCrop~\cite{zhang2025mllms}
& 57.2$^{\ddagger}$ \\
\quad + SD-RPN~\cite{shi2026sdrpn}
& \textbf{58.8}$^{\ddagger}$ \\
\quad + Q-Zoom~\cite{shi2026qzoom}
& 58.4$^{\ddagger}$ \\

\midrule
\multicolumn{2}{c}{Efficient global + local encoding} \\
\midrule
FastVLM-1.5B (reported)~\cite{fastvlm}
& 71.20$^{\dagger}$ \\
FastVLM-1.5B (reproduced)
& 71.95 \\
\quad + FAVE (Mean)          & 73.27 \\
\textbf{\quad + FAVE ($K{=}16$, $M{=}1$)}
                             & \textbf{73.55} \\

\bottomrule
\end{tabular*}

\caption{
\textbf{TextVQA accuracy.}
FAVE improves the reproduced FastVLM-1.5B baseline from 71.95 to 73.55 while retaining its efficient global pathway and adding sparse local visual evidence.
$\dagger$ denotes externally reported results.
$\ddagger$ denotes results under the unified LLaVA-1.5-7B evaluation reported by Q-Zoom~\cite{shi2026qzoom}.
}
\vspace{-5mm}
\label{tab:textvqa_comparison}
\end{table}

\subsection{Text-Rich Visual Understanding}

TextVQA~\cite{singh2019towards} provides a natural test of the global--local design because answers often depend on small, localized text regions. Such regions can be narrow or highly non-square, making them vulnerable to global downsampling or distortion under fixed-resolution resizing. Preserving their local geometry and visual detail is therefore important for reliable text understanding. We use DBNet++~\cite{DBnet} as a task-specific external proposal mechanism and retain up to $K=16$ source-image regions. Stage~1 jointly trains the FAVE ViT and reducer with question-conditioned contrastive supervision, producing one compact token per region. Stage~2 freezes the ViT, fine-tunes the reducer, and trains the local projector with a teacher-forced autoregressive answer objective. The FastVLM global pathway and Qwen2 remain frozen. Full training details are provided in Appendix D.2.

Table~\ref{tab:textvqa_comparison} shows that FAVE improves the reproduced FastVLM-1.5B baseline by \textbf{1.60 percentage points} with at most 16 additional local tokens. The model adds only 7.04M vision-side parameters and does not modify the global pathway or language model. The gain therefore comes from supplying additional local visual evidence to the existing multimodal system.

\begin{table}[t]
\centering
\small

\begin{tabular*}{\columnwidth}{@{\extracolsep{\fill}}lrr@{}}
\toprule
Method
& \shortstack{Vision enc.\\(ms) $\downarrow$}
& \shortstack{TTFT\\(ms) $\downarrow$} \\
\midrule

\multicolumn{3}{c}{Compact VLM} \\
\midrule
SmolVLM2-2.2B~\cite{marafioti2025smolvlm}
& 169.65
& 246.54 \\

\midrule
\multicolumn{3}{c}{Efficient global + local encoding} \\
\midrule
FastVLM-1.5B~\cite{fastvlm}
& 29.54
& 59.00 \\

\quad + FAVE (Mean)
& $29.54 + \textbf{9.25} = 38.79$
& 68.03 \\

\quad + FAVE ($K{=}16$, $M{=}1$)
& $29.54 + \textbf{9.18} = 38.72$
& 74.51 \\

\bottomrule
\end{tabular*}

\caption{
\textbf{TextVQA inference efficiency.}
We report vision-encoder latency and end-to-end time-to-first-token (TTFT).
For FAVE, vision latency is decomposed into frozen FastViTHD global encoder and local FAVE encoder over all selected crops.
FAVE uses a 224-pixel crop cap with Patch-n-Pack inference.
TTFT is measured on an NVIDIA A40 with batch size 1 over a fixed 500-example subset; vision latency is measured separately on 50 examples.
}
\label{tab:textvqa_latency}
\end{table}

Table~\ref{tab:textvqa_latency} shows that adding the FAVE local pathway costs 9.18\,ms of visual encoding. Detailed latency accounting is provided in Appendix D.4.

\begin{table}[t]
\centering
\small
\setlength{\tabcolsep}{4pt}
\begin{tabular}{lccc}
\toprule
Subset & FastVLM & FAVE & $\Delta$ \\
\midrule
OCRBench overall & 54.5 & \textbf{55.8} & +1.3 \\
Scene Text VQA & 66.5 & \textbf{80.0} & +13.5 \\
\midrule
ESTVQA & 58.0 & \textbf{80.0} & +22.0 \\
STVQA & 40.0 & \textbf{66.0} & +26.0 \\
OCRVQA & 94.0 & \textbf{96.0} & +2.0 \\
TextVQA & 74.0 & \textbf{78.0} & +4.0 \\
\bottomrule
\end{tabular}
\caption{
\textbf{OCRBench transfer without OCRBench-specific training.}
Values are accuracy (\%), and $\Delta$ denotes percentage-point improvement. FAVE improves the overall OCRBench score and yields its largest gains on Scene Text-centric VQA, where localized scene text is most relevant.
}
\vspace{-4mm}
\label{tab:ocrbench}
\end{table}

As a transfer test, we evaluate the same TextVQA-trained model on OCRBench~\cite{Liu2024} without OCRBench-specific training. Table~\ref{tab:ocrbench} shows a \textbf{1.3-point} overall gain and a \textbf{13.5-point} gain on Scene Text-centric VQA, with the strongest improvements on ESTVQA and STVQA. This suggests that the fine-grained local representation transfers well across related text-centric domains without task-specific adaptation.

\subsection{Fine-Grained Attribute Understanding}

We next ask whether the benefit extends beyond text to fine-grained object appearance. We therefore evaluate only questions labeled as \emph{attribute} under the GQA semantic question type~\cite{Hudson2019CVPR}. These questions probe properties such as color, material, size, state, and shape, providing a targeted test of local visual evidence rather than aggregate GQA reasoning.

A question-independent YOLOE-26S~\cite{wang2025yoloe} proposal network provides candidate object regions. We otherwise follow the TextVQA multimodal setup, except that Stage~1 is trained with supervised object and attribute semantics to preserve fine-grained appearance information. Full supervision and proposal details are provided in Appendix D.5.

\begin{table}[t]
\centering
\small

\begin{tabular*}{\columnwidth}{@{\extracolsep{\fill}}lrr@{}}
\toprule
Method
& \shortstack{Total\\params.}
& \shortstack{GQA\\Attribute $\uparrow$} \\
\midrule

\multicolumn{3}{c}{Compact VLM} \\
\midrule
SmolVLM2-2.2B~\cite{marafioti2025smolvlm}
& 2.247B
& 57.40 \\

TinyLLaVA-Phi2-SigLIP~\cite{tinyllava}
& 3.1B
& 68.40 \\

\midrule
\multicolumn{3}{c}{Fine-grained VLM} \\
\midrule
AttWarp (LLaVA-1.5-7B)~\cite{dalal2025constructive}
& $\sim$7B
& 69.30 \\
ViCrop (LLaVA-1.5-7B)~\cite{zhang2025mllms}
& $\sim$7B
& 68.20 \\
Mini-Gemini-HD-8B~\cite{MiniGemini}
& $\sim$8B
& 72.02 \\

\midrule
\multicolumn{3}{c}{FastVLM and FAVE} \\
\midrule
FastVLM-1.5B~\cite{fastvlm}
& 1.675B
& 70.73 \\

\quad + FAVE ($K{=}16$, $M{=}1$)
& 1.682B
& \textbf{72.04} \\

FastVLM-7B~\cite{fastvlm}
& $\sim$8.0B
& 72.72 \\

\bottomrule
\end{tabular*}

\caption{
\textbf{Fine-grained attribute understanding on GQA.}
We report accuracy on questions labeled as \emph{attribute} under the GQA semantic question type. FAVE improves FastVLM-1.5B from 70.73 to 72.04 while keeping the global visual pathway and language model frozen. It matches the substantially larger Mini-Gemini-HD-8B (72.02) and remains within 0.68 points of FastVLM-7B.
}
\vspace{-6mm}
\label{tab:gqa_attribute}
\end{table}

Table~\ref{tab:gqa_attribute} shows that FAVE improves FastVLM-1.5B by \textbf{1.31 percentage points}, outperforming the compact VLM baselines and matching the substantially larger multi-encoder Mini-Gemini-HD-8B (72.04 vs.\ 72.02). It also narrows the gap to FastVLM-7B to only 0.68 points while leaving the global FastVLM pathway and 1.5B language model unchanged. Together with the ImageNet and TextVQA results, this shows that targeted local visual capacity provides a complementary path to fine-grained accuracy gains without scaling the entire visual--language system.
\section{Conclusion}

Fine-grained visual evidence does not require uniformly increasing resolution or capacity across the entire image. We introduced FAVE, a foveated adaptive visual encoder that preserves native region geometry and lets local resolution, token count, and compute scale with region size. Controlled ImageNet experiments establish three effects: FAVE encodes the same selected small-object region more accurately and efficiently than fixed-resolution and variable-resolution controls; additional compute is more effective when directed to the selected region than spread across the full image; and lightweight FAVE-T reaches an accuracy--efficiency operating point that substantially larger fixed-resolution ViTs do not recover. We then pair FAVE with FastVLM as a sparse local complement to its compact global representation. With at most 16 local tokens, FAVE improves TextVQA by 1.60 percentage points while keeping FastVLM and Qwen2 frozen. On GQA attributes, it improves FastVLM-1.5B by 1.31 points, matches Mini-Gemini-HD-8B, and comes within 0.68 points of FastVLM-7B. Together, these results show that targeted local visual capacity can recover fine-grained evidence efficiently without scaling the entire visual--language system.

\section*{Acknowledgments}

This work was supported in part by the Center for the Co-Design of Cognitive Systems (CoCoSys), a DARPA-sponsored JUMP 2.0 center, the Semiconductor Research Corporation (SRC), and the National Science Foundation (NSF). We also thank Deepak Ravikumar for helpful initial discussions.
\appendix

\section{Additional Related Work}
\label{supp:related}

\subsection{Foveated Vision}

Foveated vision reduces visual processing cost by allocating
different levels of spatial detail across the visual field.
Computational models have implemented this principle through
eccentricity-dependent blur, pooling, and non-uniform sampling
inspired by human peripheral vision
\cite{akbas2014object,deza2016peripheral,deza2021emergent,shah2023training,Poggiocortical}.
These methods preserve greater visual detail near selected locations
while representing peripheral regions more coarsely.

FocL takes an object-centric approach by processing selected crops
to reduce background clutter, but uses a fixed crop resolution
\cite{mukherjee2026from}.
Segment This Thing instead applies a fixed point-centered foveation
pattern in Vision Transformers, with fine-resolution tokens near the
prompt and progressively coarser tokens toward the periphery
\cite{schmidt2025segment}.

FAVE differs in how the selected region is encoded.
Given an externally selected region, FAVE preserves its native
geometry rather than imposing a fixed-resolution representation or a
fixation-centered resolution profile.
Its spatial resolution, patch grid, token count, and visual compute
therefore vary with region scale.
Spatial selection remains separate from local encoding, allowing
FAVE to focus on how a selected region should be represented under a
bounded visual-compute budget.

\subsection{Efficient and Adaptive Visual Encoding}

Efficient visual encoders reduce computation through architectural
design, flexible input resolution, or visual-token reduction.
Swin Transformer restricts self-attention to shifted local windows,
while Focal Transformer and FocalNet aggregate information across
multiple spatial scales
\cite{SwinV2,yang2021focal,yang2022focal}.
EfficientViT targets efficient inference through a lightweight
attention design, while FastViT uses a reparameterized hybrid
architecture to improve the latency--accuracy trade-off
\cite{liu2023efficientvit,vasufastvit2023}.

Other methods make visual encoders more flexible to input resolution.
ResFormer uses multi-resolution training to improve performance
across different image sizes, while ViTAR supports high-resolution
inputs by adaptively merging visual tokens
\cite{tian2022resformer,fan2024vitar}.
A separate class of methods reduces an existing visual-token
sequence.
DynamicViT dynamically prunes tokens, Adaptive Token Sampling
selects a subset of informative tokens, and Token Merging combines
similar tokens during inference
\cite{rao2021dynamicvit,Fayyaz2021AdaptiveTS,bolya2022tome}.

NaViT instead preserves native image resolution and aspect ratio.
Patch-n-Pack allows variable-length image sequences to be processed
efficiently within a shared token budget
\cite{navit}.
FAVE uses this variable-resolution capability at the level of
externally selected regions rather than the complete image.
The native geometry of each selected region determines its patch
grid and resulting visual-token count.
FAVE therefore adapts the local spatial budget without pruning,
sampling, or merging tokens from an already constructed visual
representation.

\subsection{Efficient and Adaptive Multimodal Perception}

Efficient VLMs reduce multimodal cost at different parts of the
system. SmolVLM2 and TinyLLaVA target compact multimodal models
through smaller model configurations and efficient training and
representation choices \cite{marafioti2025smolvlm,tinyllava}.
FastVLM instead focuses directly on global visual encoding.
FastViTHD reduces vision latency and produces a compact visual-token
sequence for the language model \cite{fastvlm}.
FAVE is complementary to this approach. We retain FastVLM as the
global pathway and add visual capacity only through a lightweight
local encoder for selected regions. The original global pathway and
language model remain frozen.

Visual-token reduction provides another route to efficiency.
TokenPacker condenses high-resolution visual features through a
coarse-to-fine visual projector, while VisionZip selects informative
tokens from the representation produced by the visual encoder
\cite{li2025tokenpacker,yang2025visionzip}.
These methods reduce redundancy in an existing visual
representation. FAVE instead re-encodes selected regions directly
from the source image. It can therefore introduce local evidence that
may have been weakened before or during compact global encoding.
The resulting region representations are then compressed into a
small local-token budget before language-model inference.

High-resolution and region-adaptive methods address fine-grained
evidence in different ways. LLaVA-UHD divides the native-resolution
image into adaptive slices and compresses their features while
preserving their spatial organization \cite{guo2024llavauhd}.
AttWarp and ViCrop instead use signals from the MLLM itself to
redistribute or revisit visual detail at inference time without
training new model weights
\cite{dalal2025constructive,zhang2025mllms}.
SD-RPN learns a lightweight region predictor from pseudo-RoI labels
distilled from the MLLM's internal attention, while Q-Zoom adds
query-aware routing so that high-resolution regional processing is
invoked only when the coarse representation is insufficient
\cite{shi2026sdrpn,shi2026qzoom}.
These approaches primarily address where and when additional visual
detail should be acquired. FAVE separates this decision from local
encoding. An external proposal mechanism determines where to look,
while FAVE determines how each selected region is represented with
native geometry and an adaptive spatial budget.

Localized perception can also be integrated directly into reasoning.
Visual CoT uses intermediate localized regions in a multi-turn
reasoning pipeline and provides training data with region and
reasoning annotations \cite{shao2024visual}.
Foveated Reasoner makes high-acuity visual acquisition an action
within the autoregressive reasoning trajectory and trains this
behavior using supervised initialization followed by reinforcement
learning \cite{min2026foveatedreasoning}.
FAVE does not place visual acquisition inside the language-model
reasoning loop. Selected regions are encoded before generation, and
their local tokens augment the visual prefix consumed by the frozen
language model. LLMind provides another complementary design. It
uses training-free, bio-inspired non-uniform image sampling under a
constrained pixel budget rather than adding a separately learned
local encoder \cite{debnath2026llmind}.

Multi-encoder VLMs combine complementary visual representations
within one multimodal architecture. Mini-Gemini adds a second
high-resolution visual encoder and uses its features to refine the
visual representation without increasing the final visual-token
count \cite{MiniGemini}. Cambrian-1 studies combinations of multiple
vision encoders and uses a spatially aware aggregator to integrate
high-resolution features with the language model
\cite{tong2024cambrian}.
Our VLM instantiation also uses complementary visual pathways, but
with a different division of labor. FastVLM retains the efficient
global representation, while FAVE independently re-encodes
externally selected regions. Spatial selection remains separate from
local encoding, the local representations are compressed into a
bounded token budget, and the global FastVLM pathway and language
model remain frozen.


\section{Additional FAVE Implementation Details}
\label{sec:supp_fave_impl}

This section provides the additional FAVE implementation and training
details referenced in the main paper. ImageNet dataset construction and
evaluation protocols are described separately in Sec.~C.

\subsection{Object-Centric Crop Construction}

For ImageNet training, FAVE processes one selected object-centric crop per
sample. We perturb the selected box during training using an isotropic scale
factor sampled from $[0.9,1.1]$. We also perturb the crop center independently
along the horizontal and vertical axes by at most $20\%$ of the corresponding
box dimension. A minimum crop side of 32 pixels is enforced. Evaluation uses
the deterministic GT-selected crop without spatial jitter.

FAVE does not resize every selected region to $224\times224$. Instead, the
maximum resolution is chosen according to the operating resolution of the
task. For ImageNet, we use 224 pixels because this is the resolution used by
the standard fixed-resolution setup and by the corresponding baselines in our
comparison. Crops whose longest side is at most 224 pixels therefore remain at
native scale. Only larger crops are isotropically downsampled so that their
longest side is 224 pixels. If a task used a different reference resolution,
the same procedure would use that resolution as the task-specific cap. Thus,
224 is an ImageNet-specific operating point rather than a fixed property of
FAVE.

Height and width are never resized independently. The resulting crop is
aligned to the 16-pixel patch grid without anisotropic warping, following the
grid-alignment procedure described in the main paper. The spatial dimensions
presented to the encoder therefore remain crop dependent rather than being
normalized to a common $224\times224$ input.

\subsection{Variable-Resolution FAVE Encoder}

The ImageNet FAVE-T encoder follows the Tiny-scale configuration used in the
main comparison. It contains 12 Transformer blocks with embedding dimension
$D=192$, three attention heads, an MLP expansion ratio of 4, and
$16\times16$ patch embedding. Patch embedding is implemented with kernel size
and stride 16. A learned CLS token is prepended to each crop sequence.

Because crop dimensions remain adaptive, different samples produce different
numbers of patch tokens. All valid patch tokens are retained and processed by
the same encoder. The final CLS representation is normalized and passed to a
shared linear ImageNet classifier. We use stochastic depth increasing linearly
to 0.1 across the Transformer blocks.

\subsection{Variable-Length Batching and Patch-n-Pack}

We train the ImageNet FAVE model using standard padded variable-length
sequence batching. Each crop first forms its complete sequence
$[\mathrm{CLS}_i \mid x_{i,1},\ldots,x_{i,P_i}]$. Within a batch, sequences
are padded to the longest sequence, and padding positions are masked from
attention and from the classification objective.

A model trained using this standard sequence-batching procedure can be used
directly with Patch-n-Pack at inference time; no retraining or parameter
modification is required. Patch-n-Pack changes only how the already formed
per-image sequences are scheduled. Multiple complete sequences are
concatenated into a packed row under a token budget $T_{\max}$, while a
block-diagonal attention mask prevents interaction between different images.
Each image retains its own CLS token and all of its patch tokens.

The per-image computation on valid tokens is therefore unchanged between the
two execution schemes. Crop construction, patch embedding, encoder weights,
and classifier weights are identical. Patch-n-Pack changes batching and
sequence scheduling rather than the learned prediction function, and it does
not change the per-image token count or the analytical FLOP accounting
reported in the main paper.

\subsection{Optimization}

FAVE-T is trained for 300 epochs with AdamW. The training batch size is 64
images per GPU, and the learning rate is linearly scaled with the global batch
size from a base learning rate of $10^{-3}$ at batch size 512. We use weight
decay 0.05, Adam coefficients $(0.9,0.999)$, and gradient clipping at norm
1.0. The learning rate is warmed up for 10 epochs and then follows a cosine
schedule to a minimum value of $10^{-6}$. Training uses cross-entropy with
label smoothing 0.1.

Bias terms, normalization parameters, positional embeddings, and the CLS
token are excluded from weight decay. We maintain an exponential moving
average of the model parameters with decay 0.99996 after a two-epoch warm-up.
Training uses random horizontal flipping and photometric augmentation,
followed by ImageNet normalization and size-aware random erasing. Validation
uses deterministic crop construction and ImageNet normalization.

The matched Adaptive ViT-T full-image control uses the same optimization,
regularization, EMA, and appearance-augmentation settings as FAVE-T, differing
primarily in its spatial input construction. The fixed-resolution ViT-S/16
and ViT-B/16 capacity controls are trained from scratch for 300 epochs on the
same ImageNet train--validation partition using strong fixed-resolution ViT
training recipes. In particular, these larger models use RandomResizedCrop,
Mixup/CutMix, label smoothing, and random erasing, together with
architecture-specific batch sizes, learning-rate scaling, weight decay, and
warm-up schedules. Thus, the capacity controls are trained with optimization
and regularization appropriate to their model scale rather than being forced
to share FAVE-T's hyperparameters.

\section{Additional ImageNet Experimental Details}
\label{sec:supp_imagenet_details}

\subsection{Dataset Split and Annotation Protocol}

We use the ImageNet subset with available bounding-box annotations. Following
the supervised ImageNet split used in FocL~\cite{mukherjee2026from}, we use a
fixed 90/10 train--validation partition. The same partition is used for FAVE
and all ImageNet controls reported in the main paper, so differences between
methods are not caused by differences in training data.

Only samples with at least one valid ground-truth bounding box are retained.
When an image contains multiple valid boxes, we use the largest box by area as
the selected object. This convention is used consistently for object-centric
crop construction and for defining the native object scale used in the
evaluation cohorts.

\subsection{Evaluation Cohorts}

We evaluate on three independently sampled 2,000-image validation cohorts,
defined using the native maximum side
\(S=\max(w,h)\) of the largest GT bounding box. The first cohort contains
images with \(S\leq96\) pixels, the second contains images with
\(S\leq224\) pixels, and the General-2K cohort is sampled without an object-size
restriction. Each cohort is sampled independently after applying its
corresponding eligibility rule. They should therefore be interpreted as
filtered evaluation cohorts rather than as disjoint object-size bins or nested
2K subsets.

The \(96\)-pixel threshold corresponds to the first two smallest-scale bins
of the bounding-box scale partition used to characterize the dataset:
\(S\leq48\) and \(49\leq S\leq96\). We therefore use their union as the
stringent small-object cohort. The \(224\)-pixel cohort provides a broader
object-scale setting while remaining aligned with the ImageNet operating
resolution used in this study.

\subsection{Evaluation and Resolution Policies}

For object-centric evaluation, the adaptive models use the same deterministic
GT-selected region. Fixed-Res ViT-T/16 receives the same selected crop window,
but resizes it to $224\times224$ before encoding. Adaptive ViT-T (Full) and
FAVE instead preserve the adaptive crop geometry described in Sec.~B. For
full-image evaluation, Adaptive ViT-T (Full) preserves aspect ratio and
downsamples only when the image's longest side exceeds 224 pixels.

For the adaptive-zoom evaluation in Table~2 of the main paper, the
magnification factor is selected from the native maximum side $S$ of the GT
bounding box. We use $1.8\times$ for $S<96$, $1.5\times$ for
$96\leq S<128$, $1.4\times$ for $128\leq S<144$, and $1.2\times$ otherwise.
Magnification is applied to the selected region before the same 224-pixel
task-specific cap and patch-grid alignment used for native FAVE inference.
NaViT-T's long-side-cap policy preserves the complete image and downsamples
only when its longest side exceeds 224 pixels. Its $r=224$ policy instead
targets an effective image area of approximately $224^2$ pixels while
preserving aspect ratio, and can therefore either upsample or downsample the
complete image.

\subsection{Token and Analytical FLOP Accounting}

We report visual-token count and analytical forward FLOPs at the actual
spatial resolution processed for each image. For the CLS-based models, an
input with patch-grid dimensions $H_p\times W_p$ contains
$P=H_pW_p$ patch tokens and has sequence length $N=P+1$. Token counts are
computed independently for every image and averaged over the corresponding
2,000-image cohort. They therefore reflect valid per-image sequence lengths
rather than padded batch lengths. NaViT-T uses learned-query pooling rather
than a CLS token and consequently reports patch tokens only.

For the CLS-based ViT models, the dominant Transformer-block cost is computed
as
\[
L\left(12ND^2+2N^2D\right),
\]
for embedding dimension $D$ and depth $L$. We additionally account for patch
projection and the classification head using the same fixed analytical
convention for all reported operating points. Layer normalization, softmax,
nonlinearities, bias additions, and runtime padding overhead are excluded. We
count a multiplication and an addition as two floating-point operations.
Because self-attention is quadratic in sequence length, FLOPs are computed
independently from each image's actual valid sequence length and then averaged
over the cohort, rather than by substituting the mean token count into the FLOP
expression. Reported values therefore describe logical per-image model compute
on valid tokens and exclude runtime work introduced by padding or packed
execution. NaViT-T is accounted for analogously, with the additional cost of
its learned-query attention pooling.

As a reference, a fixed $224\times224$ ViT-T/16 processes 196 patch tokens
plus one CLS token and requires 1.283 analytical GFLOPs per image under this
accounting. On the $\leq96$ cohort, FAVE processes 17.7 visual tokens and
0.101 analytical GFLOPs per image on average, corresponding to the
$12.7\times$ reduction reported in the main paper. Patch-n-Pack changes
sequence scheduling but not the per-image token count or analytical FLOP
accounting.

\section{Additional Multimodal Experimental Details}
\label{sec:supp_vlm_arch}
\subsection{Common Multimodal FAVE Architecture}

We use the common global--local architecture described in
Sec.~4 and Fig.~3 of the main paper. The complete image is processed by the
frozen FastVLM vision tower and multimodal projector, producing 256 global
visual tokens. In parallel, a task-specific external proposal mechanism
selects up to \(K\) regions. Importantly, these regions are cropped directly
from the original source image, rather than from the resized input or
intermediate representation used by the FastVLM global pathway. Each selected
source-image crop is then independently processed by FAVE using the
variable-resolution formulation of Sec.~3, preserving its local geometry.
A shared learned cross-attention reducer compresses the resulting
variable-length patch representation to one compact token per crop. Thus, the
spatial resolution and number of FAVE patch features may vary across regions,
while each selected region contributes a fixed one-token representation to the
subsequent multimodal stage.

Training follows the two-stage procedure in Fig.~3. In Stage~1, the FAVE ViT
and token reducer are trained jointly from scratch to learn the compact local
representation. The architecture is shared across the multimodal settings,
but the proposal source and supervision used to learn this representation are
task-specific and are described in the corresponding dataset sections below.
In Stage~2, the FAVE ViT is frozen, the reducer is fine-tuned, and a local
projector trained from scratch maps each compact crop representation into the
Qwen2 embedding space. The projected local tokens are concatenated with the
256 FastVLM global tokens and the task prompt for answer generation. The
FastVLM vision tower, global multimodal projector, and Qwen2 language model
remain frozen throughout these experiments.

\subsection{TextVQA Training and Evaluation Details}
\label{sec:supp_textvqa}

\paragraph{Region proposals and supervision sources.}
For TextVQA, DBNet++ provides the local regions processed by FAVE during
Stage 2 and inference. These regions are cropped directly from the original
source image, independently of the preprocessing used by the FastVLM global
pathway. Rosetta plays a different role. During Stage~1, its OCR boxes provide
cleaner text-region geometry from which question-conditioned answer relevance
can be estimated. Rosetta therefore provides weak relevance supervision,
whereas DBNet represents the proposal distribution encountered by the
downstream model.

We transfer this supervision from Rosetta to DBNet using spatial coverage.
For a DBNet proposal \(B_{\mathrm{DB}}\) and an answer-positive Rosetta box
\(B_{\mathrm{R}}\), we measure
\begin{equation}
\frac{|B_{\mathrm{DB}}\cap B_{\mathrm{R}}|}
{|B_{\mathrm{R}}|}.
\end{equation}
A DBNet proposal is considered positive when this coverage is at least
\(0.5\), while coverage in \([0.2,0.5)\) is treated as ambiguous and excluded
from the relevance loss. We use coverage rather than IoU because a DBNet text
line can correctly contain a substantially smaller answer word while having
low IoU. For cross-source feature alignment, we further require
\begin{equation}
\frac{|B_{\mathrm{DB}}\cap B_{\mathrm{R}}|}
{|B_{\mathrm{DB}}|}
\geq 0.3 ,
\end{equation}
preventing large text-line proposals that merely contain a small Rosetta word
from being forced into feature correspondence. Rosetta OCR text can
additionally appear in the TextVQA language prompt; this prompt-level role is
separate from the DBNet proposal mechanism.

\paragraph{Stage-1 local representation learning.}
The ``question-conditioned contrastive supervision'' described in the main
paper is implemented as a hybrid objective that combines two-view visual
consistency, question-conditioned region relevance, and cross-source
alignment. Stage~1 jointly trains the FAVE ViT and learned reducer using
\begin{equation}
\mathcal{L}_{\mathrm{S1}}
=
\mathcal{L}_{\mathrm{Rosetta}}
+
\lambda_{\mathrm{db}}\mathcal{L}_{\mathrm{DBNet}}
+
\lambda_{\mathrm{cross}}\mathcal{L}_{\mathrm{cross}},
\end{equation}
where
\begin{equation}
\mathcal{L}_{\mathrm{domain}}
=
\mathcal{L}_{\mathrm{cons}}
+
\lambda_{\mathrm{rel}}\mathcal{L}_{\mathrm{rel}}.
\end{equation}
We use \(\tau=0.07\), \(\lambda_{\mathrm{rel}}=0.5\),
\(\lambda_{\mathrm{db}}=1.0\), and
\(\lambda_{\mathrm{cross}}=0.1\). For \(M>1\), the consistency and relevance
terms are evaluated independently for each active reducer slot and averaged
across slots. The reported \(K=16,M=1\) configuration does not use the
optional reducer-slot diversity regularizer.

The consistency term is a two-view InfoNCE objective. Each region is rendered
twice with independent augmentation; the two views of the same region form a
positive pair, while views from all other regions are negatives, including
other regions from the same image:
\begin{equation}
\ell_i^{\mathrm{cons}}
=
-\log
\frac{
\exp\!\left(s(z_i,z_i^{+})/\tau\right)
}{
\sum_{j\neq i}
\exp\!\left(s(z_i,z_j)/\tau\right)
}.
\end{equation}
This term encourages the compact local representation to remain stable under
small appearance and crop perturbations while retaining discriminative
information across regions. Importantly, this InfoNCE term is not itself
question-conditioned.

Question-conditioned supervision enters through
\(\mathcal{L}_{\mathrm{rel}}\). The representations of the two augmented views
of each region are averaged and re-normalized. For each question, the mean
embedding of its answer-positive regions defines a normalized positive
reference, and candidate regions are trained with a weighted binary relevance
loss using their cosine similarity to this reference. Exact, numeric, and
concatenation answer matches receive weight \(1\), substring matches receive
weight \(0.6\), and matched Rosetta tokens shorter than three normalized
characters are ignored. Stage~1 therefore learns two complementary
properties: augmentation-stable local representations through two-view
contrastive learning and preservation of answer-relevant text evidence through
weak question-conditioned supervision.

Finally, confidently corresponding Rosetta and DBNet regions are aligned
through
\begin{equation}
\mathcal{L}_{\mathrm{cross}}
=
1-\cos
\left(
z_{\mathrm{Rosetta}},
z_{\mathrm{DBNet}}
\right).
\end{equation}
This transfers supervision derived from the cleaner Rosetta regions toward the
DBNet proposal domain used by the downstream model.

\paragraph{Learned crop compression.}
The FAVE ViT produces a variable number of \(192\)-D patch features for each
source-image crop. The mean-pooling baseline compresses these features as
\begin{equation}
\mathbf{z}
=
\frac{1}{N}\sum_{i=1}^{N}\mathbf{x}_i,
\end{equation}
assigning equal weight to every patch. The learned reducer instead uses shared
learned queries that cross-attend to the variable-length patch sequence and
produce \(M\) compact \(192\)-D representations. For the reported \(M=1\)
configuration, a single shared learned query produces one representation per
selected region. The ViT and reducer are optimized jointly during Stage~1,
allowing the retained token to emphasize crop evidence useful under the
Stage-1 objectives rather than uniformly averaging all patches.

During Stage~1 training, the active reducer width \(M\) is sampled from
\(\{1,2,3\}\), providing supervision to each reducer prefix. The Stage~1
checkpoint is selected using the mean validation loss across
\(M=1,2,3\). A temporary
\(192\!\rightarrow384\!\rightarrow128\) projection head is used only for the
Stage~1 objectives and is discarded afterward.

\paragraph{Stage-2 multimodal adaptation.}
Stage~2 connects the compact local representations to the frozen FastVLM
system. The FAVE ViT is frozen, the learned reducer is fine-tuned, and a
\(192\!\rightarrow512\!\rightarrow1536\) local projector together with a
lightweight auxiliary region-relevance head is trained. The projected local
representations enter the Qwen2 embedding space. For the reported
configuration, up to \(K=16\) DBNet regions are retained and \(M=1\), adding
at most 16 local visual tokens to the 256 FastVLM global tokens. The FastVLM
vision tower, global multimodal projector, Qwen2 language model, and language
head remain frozen.

Stage~2 is optimized with
\begin{equation}
\mathcal{L}_{\mathrm{S2}}
=
\mathcal{L}_{\mathrm{sharp\text{-}MA}}
+
\lambda_{\mathrm{aux}}\mathcal{L}_{\mathrm{region}},
\end{equation}
where \(\lambda_{\mathrm{aux}}=0.03\) for the reported learned-reducer
configuration. The auxiliary region loss uses Rosetta-derived weak
question-conditioned relevance labels transferred to the selected DBNet
regions used by the VLM.

TextVQA provides multiple human answers for each question. Rather than reducing
these annotations to a single hard target, we first merge answers that produce
the same supervised token sequence. If the resulting answer \(a\) occurs
\(c_a\) times, it receives weight
\begin{equation}
w_a
=
\frac{c_a^{\,\gamma}}
{\sum_b c_b^{\,\gamma}},
\qquad
\gamma=1.5 .
\end{equation}
A separate teacher-forced autoregressive cross-entropy is computed for each
distinct answer sequence and combined according to \(w_a\). Ordinary
count weighting corresponds to \(\gamma=1\). Sharpening with
\(\gamma=1.5\) retains different valid surface forms among the human
annotations while increasing the relative contribution of answers with
stronger annotator consensus.

\paragraph{Optimization and model selection.}
Stage~1 is optimized with AdamW using a common learning rate for the FAVE
ViT, reducer, and temporary projection head, with ViT-style weight-decay
exclusions for biases, normalization parameters, positional and class tokens,
and reducer query tokens. We use linear warmup followed by cosine learning-rate
decay and no EMA. Validation evaluates \(M=1,2,3\) separately, and the
Stage~1 checkpoint is selected by the lowest mean validation loss across the
three reducer widths. Because the effective Stage~1 dataset retains only
matched hybrid examples, its indexed size can be smaller than the underlying
TextVQA train/development split.

Stage~2 is trained for three epochs with AdamW. We use learning rates of
\(10^{-4}\), \(10^{-5}\), and \(10^{-4}\) for the reducer, local projector,
and auxiliary region-relevance head, respectively, with zero weight decay.
The batch size is 8 with gradient accumulation over two steps. We linearly
warm up the learning rate over the first \(3\%\) of optimizer steps and then
apply cosine decay, with gradient-norm clipping at \(1.0\). Validation is
performed after each epoch, and the checkpoint with the lowest sharpened
multi-answer cross-entropy on the development set is used for final
evaluation. Checkpoint selection therefore does not use the auxiliary region
loss or accuracy on the official TextVQA evaluation set.

\paragraph{Crop construction and evaluation.}
Stage~1 regions are denormalized to original-image pixel coordinates and
cropped directly from the source image. Crops remain variable-sized, are
aligned to the \(16\)-pixel patch grid, and use a maximum side of 384 pixels,
matching the preprocessing contract of the reducer-aware checkpoint used by
the reported model. Training uses size-dependent mild translation and scale
jitter, brightness, contrast, and saturation augmentation, and \(5\%\) random
grayscale. We do not use flips, rotations, or perspective transformations.
Very small regions additionally receive a 4-pixel context margin.

We use the 34,063/539 TextVQA train/development split for model development.
Stage~1 further retains only examples satisfying its matched-region
construction. For the reproduced FastVLM and FAVE evaluations, we follow the
OCR-conditioned LLaVA-style TextVQA evaluation protocol used by FastVLM and
instantiate the OCR text with Rosetta transcriptions. This prompt-level OCR
input is identical for the reproduced FastVLM baseline and FAVE and is
separate from the DBNet proposals used by the FAVE local pathway. Results for
other methods in the main paper are taken from their corresponding reported
evaluation settings.

We evaluate the selected Stage~2 checkpoint separately on the official
5,000-example TextVQA validation set using greedy Qwen2 decoding with
\texttt{max\_new\_tokens=16}. The final TextVQA model uses \(K=16\) and
\(M=1\), adding at most 16 local visual tokens. This configuration improves
the reproduced FastVLM-1.5B baseline from \(71.95\) to \(73.55\).

\subsection{OCRBench Transfer Protocol}
\label{sec:supp_ocrbench}

We evaluate the selected TextVQA-trained FAVE model on OCRBench without any
OCRBench-specific training or parameter updates. We retain the same DBNet-based
local pathway, \(K=16,M=1\) learned reducer, and 384-pixel crop preprocessing
used by the TextVQA model.

Our primary interest is the Scene Text-centric VQA category. This setting is
the closest match to the motivation and supervision used for TextVQA: answers
depend on localized text embedded in natural scenes, where compact global
encoding may weaken small task-relevant evidence. OCRBench also contains
document-, chart-, and other OCR-oriented categories that are not directly
targeted by our TextVQA training setup. We therefore interpret the overall
OCRBench score as a broad transfer measure, while treating Scene Text-centric
VQA as the most direct test of whether the learned local representation
transfers to related scene-text understanding.

FAVE improves the overall OCRBench score from \(54.5\%\) to \(55.8\%\), while
Scene Text-centric VQA improves from \(66.5\%\) to \(80.0\%\). The largest
subset gains occur on ESTVQA (\(58.0\%\rightarrow80.0\%\)) and STVQA
(\(40.0\%\rightarrow66.0\%\)), showing that the TextVQA-trained local
representation transfers to related scene-text domains without task-specific
adaptation.

\subsection{TextVQA Efficiency and Latency Accounting}
\label{sec:supp_textvqa_latency}

\paragraph{TTFT protocol.}
Throughout the paper, time-to-first-token (TTFT) refers to end-to-end VLM
inference latency: the timed path begins with visual encoding and continues
through multimodal token construction, language-model prefill, and generation
of the first output token. External preprocessing is performed before this
timed VLM forward path. In particular, image loading, OCR prompt construction,
and DBNet proposal lookup and crop extraction are not included in TTFT.
Operationally, the timed region is exactly one synchronized
\texttt{model.generate(max\_new\_tokens=1)} call after external input
preprocessing has completed.

We measure TTFT on an NVIDIA A40 with batch size 1 over the first 500
TextVQA examples in fixed dataset order. FastVLM and FAVE use the same
OCR-conditioned prompt and input ordering. For each example, we perform three
warm-up calls followed by five timed calls to
\texttt{generate(max\_new\_tokens=1)} using greedy decoding
(\texttt{do\_sample=False}, \texttt{num\_beams=1}). Timing uses synchronized
wall-clock measurements: CUDA is synchronized immediately before and after
each timed generation call. We report the arithmetic mean over the resulting
measurements. Normal TextVQA accuracy evaluation instead allows up to 16
generated tokens.

\paragraph{Vision-encoder accounting.}
Vision latency is measured separately on a fixed 50-example subset with five
repeats per example. The frozen FastViTHD global encoder requires
\(29.540\) ms/image. For FAVE, Table~5 of the main paper reports the visual
cost as the sum of this global-encoder latency and the local FAVE encoding
cost over all selected crops. Under the reported 224-pixel crop cap with
Patch-n-Pack inference, the local encoder requires \(9.252\) ms for mean
pooling and \(9.176\) ms for the learned \(K=16,M=1\) reducer, giving total
visual-encoder costs of \(38.792\) and \(38.716\) ms, respectively. The
corresponding VLM TTFT values are \(68.026\) and \(74.510\) ms, matching the
rounded values reported in the main paper. Vision-encoder latency is reported
as component-level accounting, whereas TTFT is measured directly over the
complete VLM inference path from visual encoding through generation of the
first output token.

\paragraph{Crop-cap sensitivity.}
The trained TextVQA models use the 384-pixel crop preprocessing described in
Sec.~\ref{sec:supp_textvqa}, whereas the controlled latency experiment in
Table~5 of the main paper uses a 224-pixel local-crop cap. We additionally
measure the corresponding 384-pixel Patch-n-Pack configuration. For mean
pooling, local FAVE latency changes from \(9.252\) to \(9.195\) ms and TTFT
from \(68.026\) to \(67.928\) ms. For the learned \(K=16,M=1\) model, local
latency changes from \(9.176\) to \(9.231\) ms and TTFT from \(74.510\) to
\(74.759\) ms. These small variations in both directions indicate that the
latency conclusions are not materially affected by the controlled crop cap.

For the compact-VLM efficiency reference in Table~5, SmolVLM2-2.2B is
evaluated in BF16 using PyTorch scaled dot-product attention (SDPA). Although
SmolVLM2 also supports FlashAttention-2, we use SDPA for this comparison
because the FastVLM implementation does not use FlashAttention-2. SmolVLM2 is evaluated on the same ordered 500 question--image pairs and with
the same synchronized one-token generation boundary; each model uses its
native chat template and tokenization outside the timed region.

\begin{figure*}[t!]
    \centering

    \includegraphics[width=0.74\textwidth]{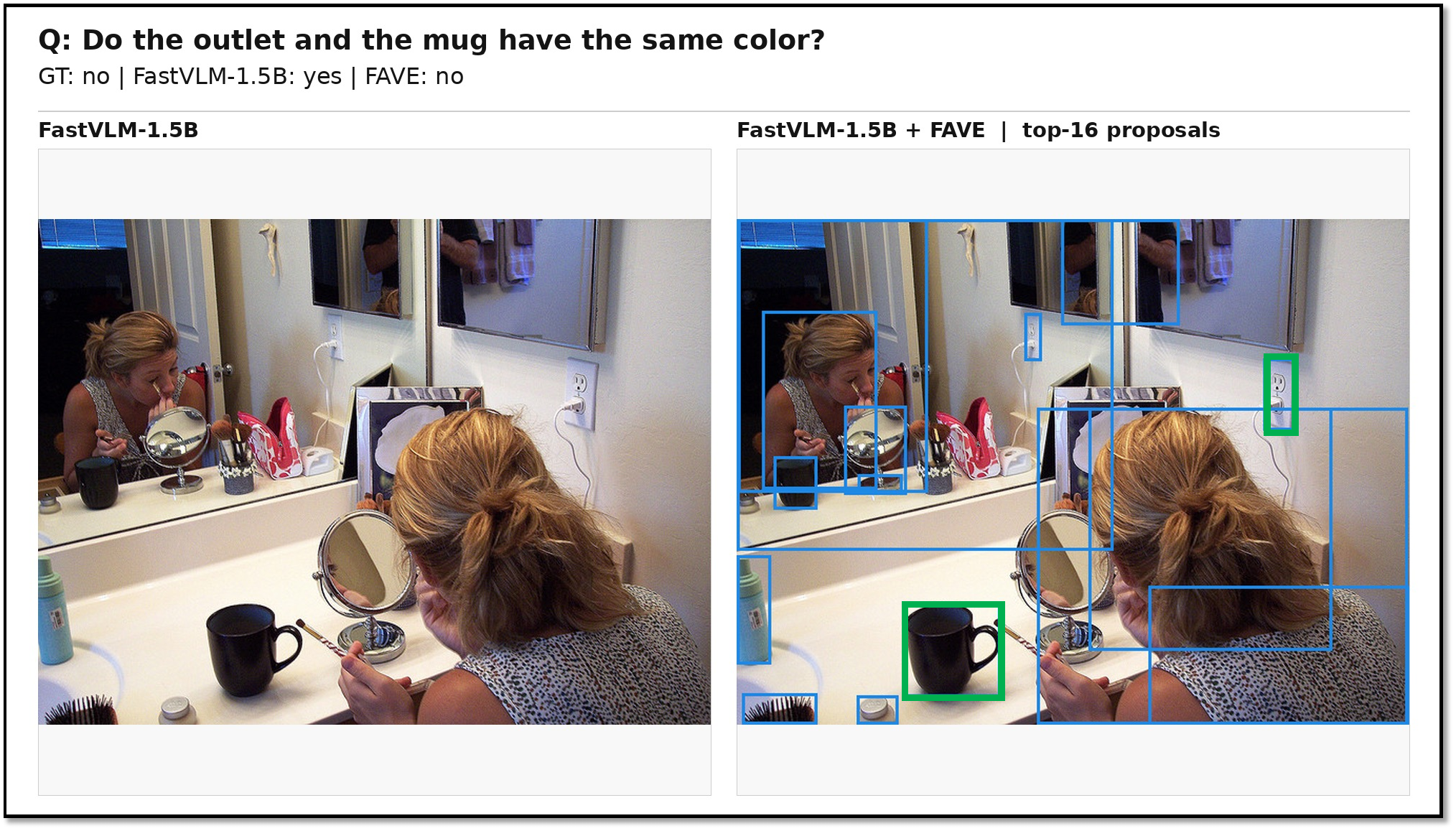}

    \smallskip

    \includegraphics[width=0.74\textwidth]{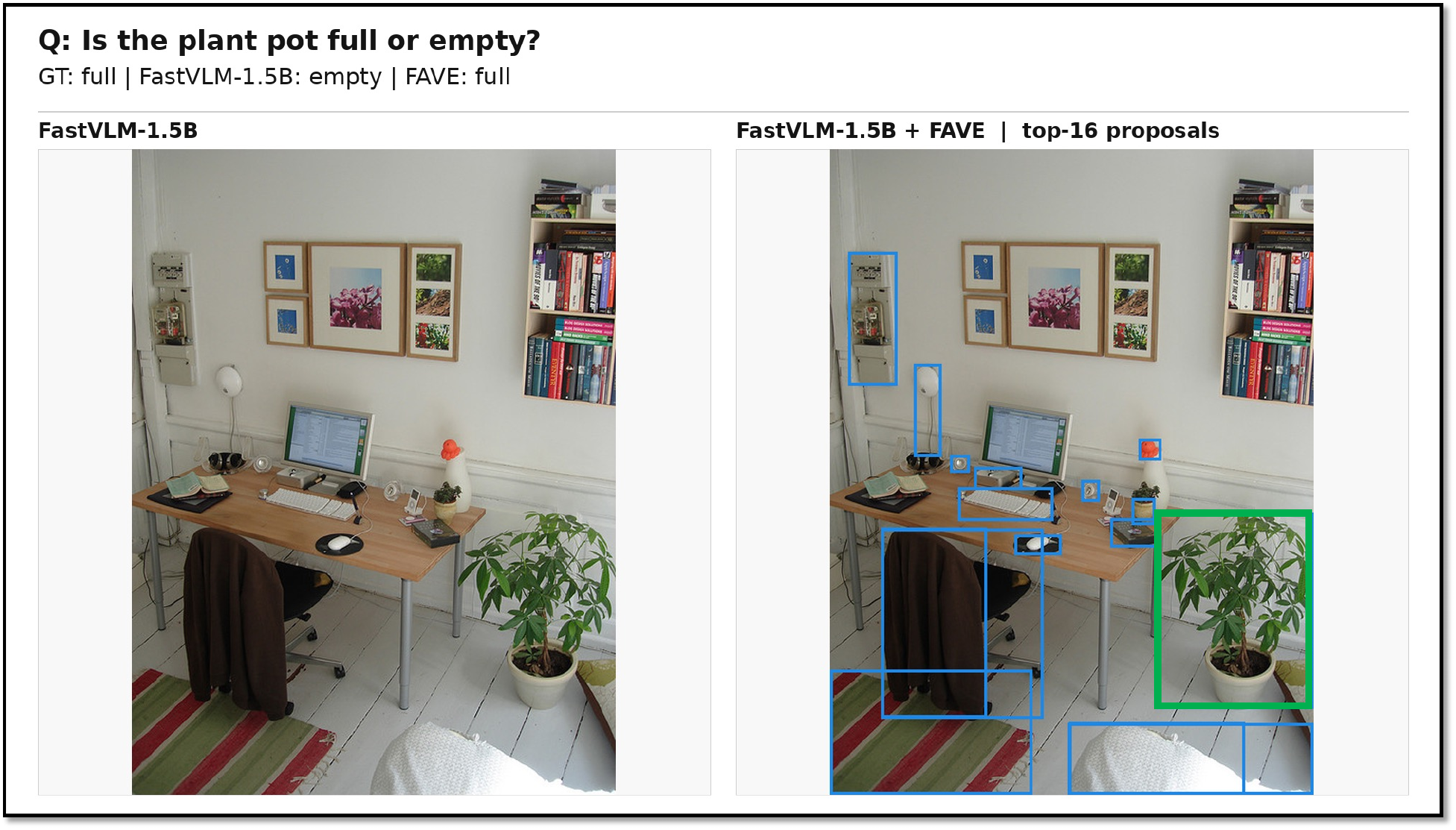}

    \caption{
    \textbf{Qualitative GQA examples of local attribute and state recognition.}
    FAVE correctly distinguishes the outlet and mug colors and recognizes that
    the plant pot is \emph{full} rather than \emph{empty}. The latter closely
    echoes the coarse-to-fine motivation of Figure~1 in the main paper:
    localized high-acuity processing recovers fine object-state evidence that
    can be weakened in the compact global representation.
    }
    \label{fig:gqa_qual_1}
\end{figure*}

\begin{figure*}[t!]
    \centering

    \includegraphics[width=0.74\textwidth]{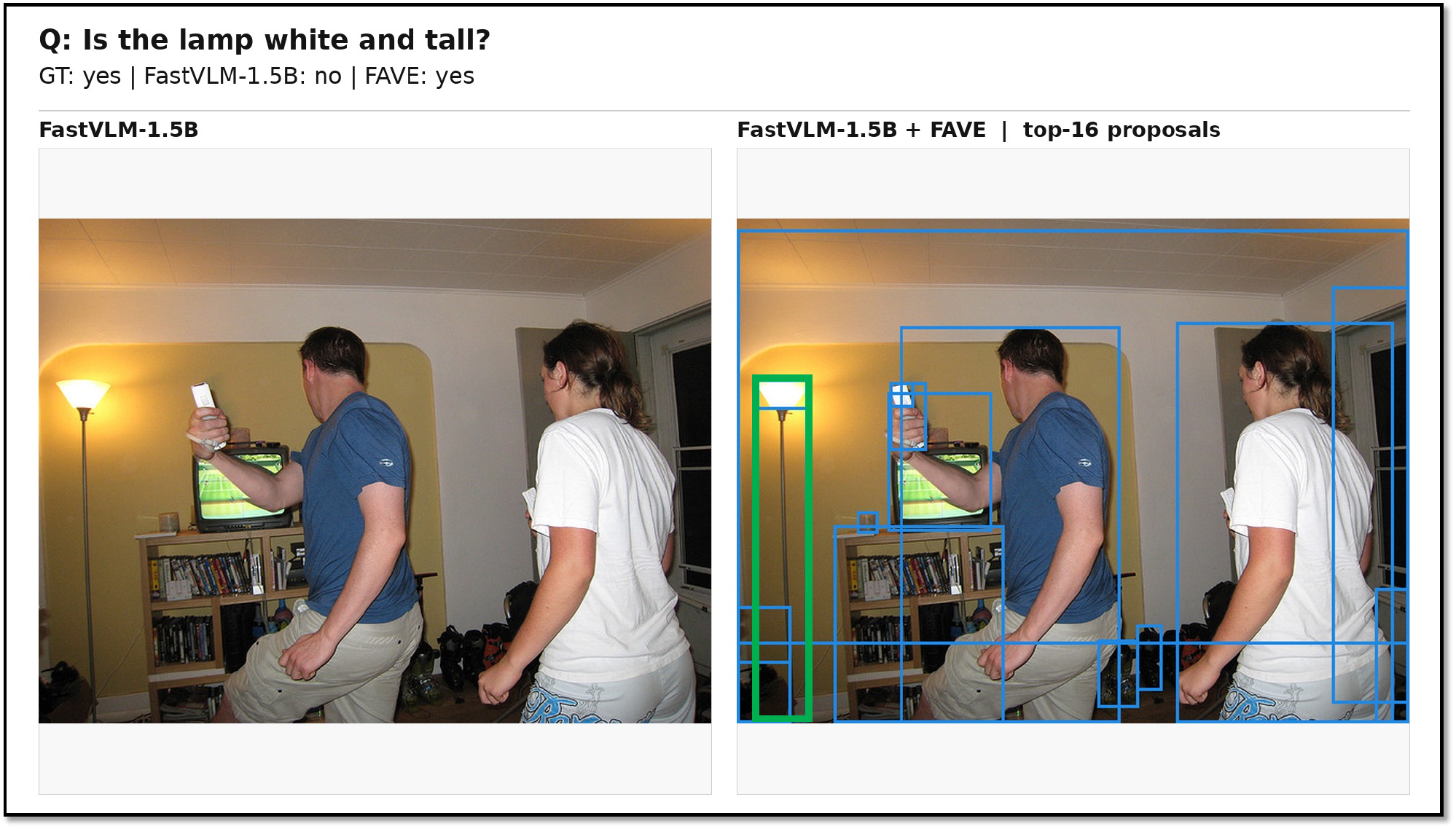}

    \smallskip

    \includegraphics[width=0.74\textwidth]{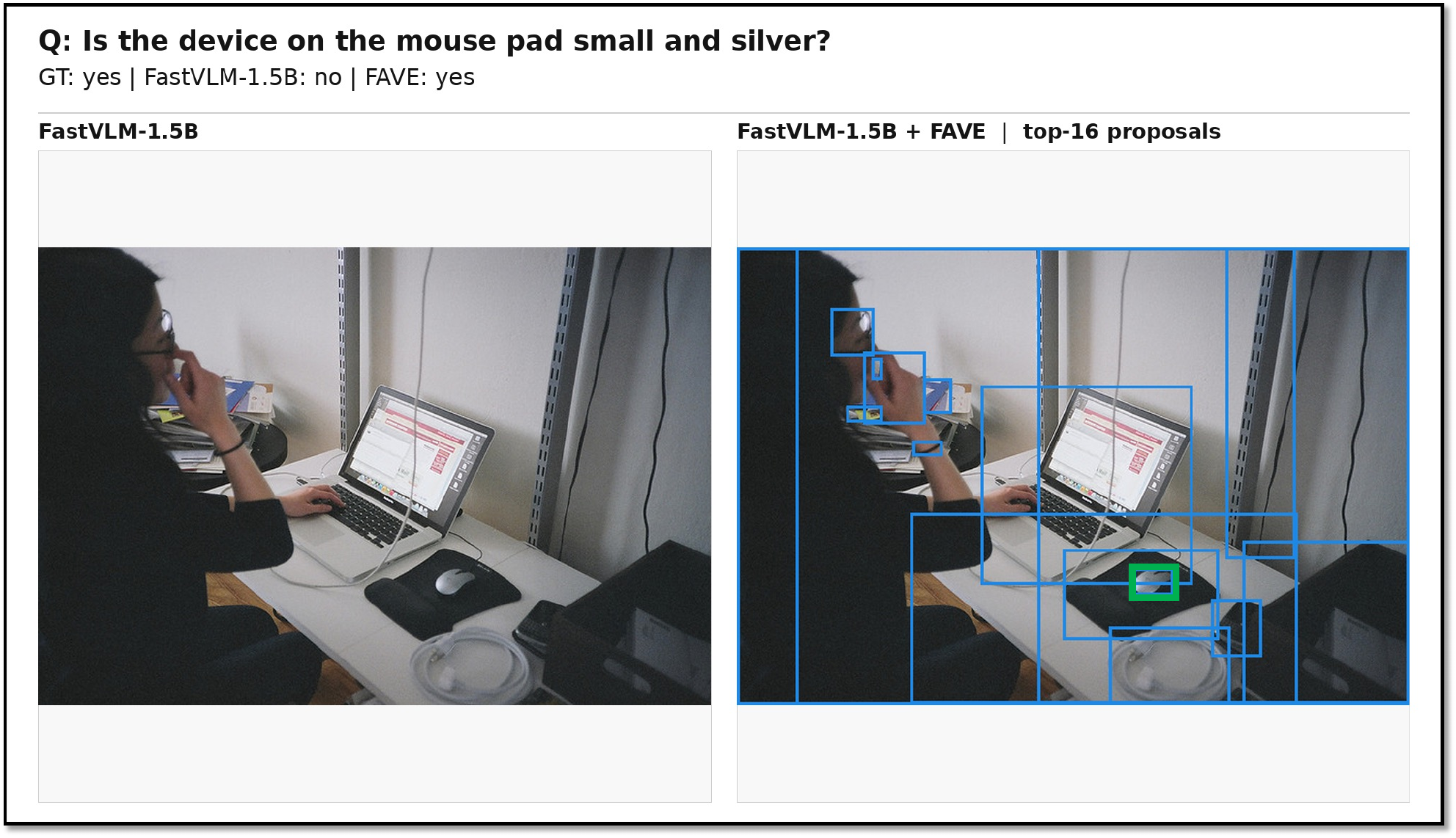}

    \caption{
    \textbf{Qualitative GQA examples of fine-grained object-property recovery.}
    FAVE correctly identifies the lamp as \emph{white} and \emph{tall} and
    the device on the mouse pad as \emph{small} and \emph{silver}, recovering
    the ground-truth answers in cases where FastVLM-1.5B fails.
    }
    \label{fig:gqa_qual_2}
\end{figure*}

\begin{figure*}[t!]
    \centering

    \includegraphics[width=0.74\textwidth]{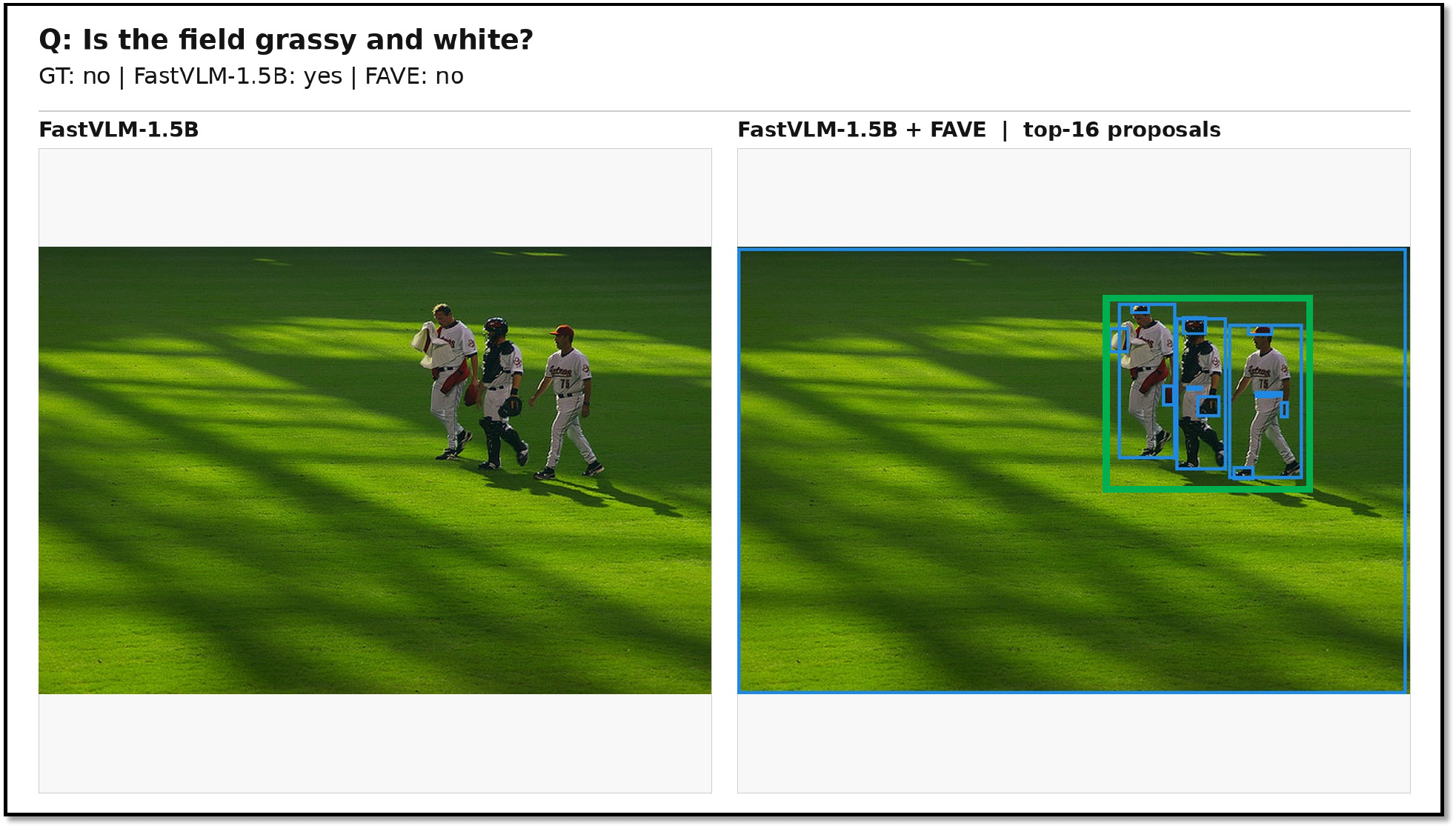}

    \smallskip

    \includegraphics[width=0.74\textwidth]{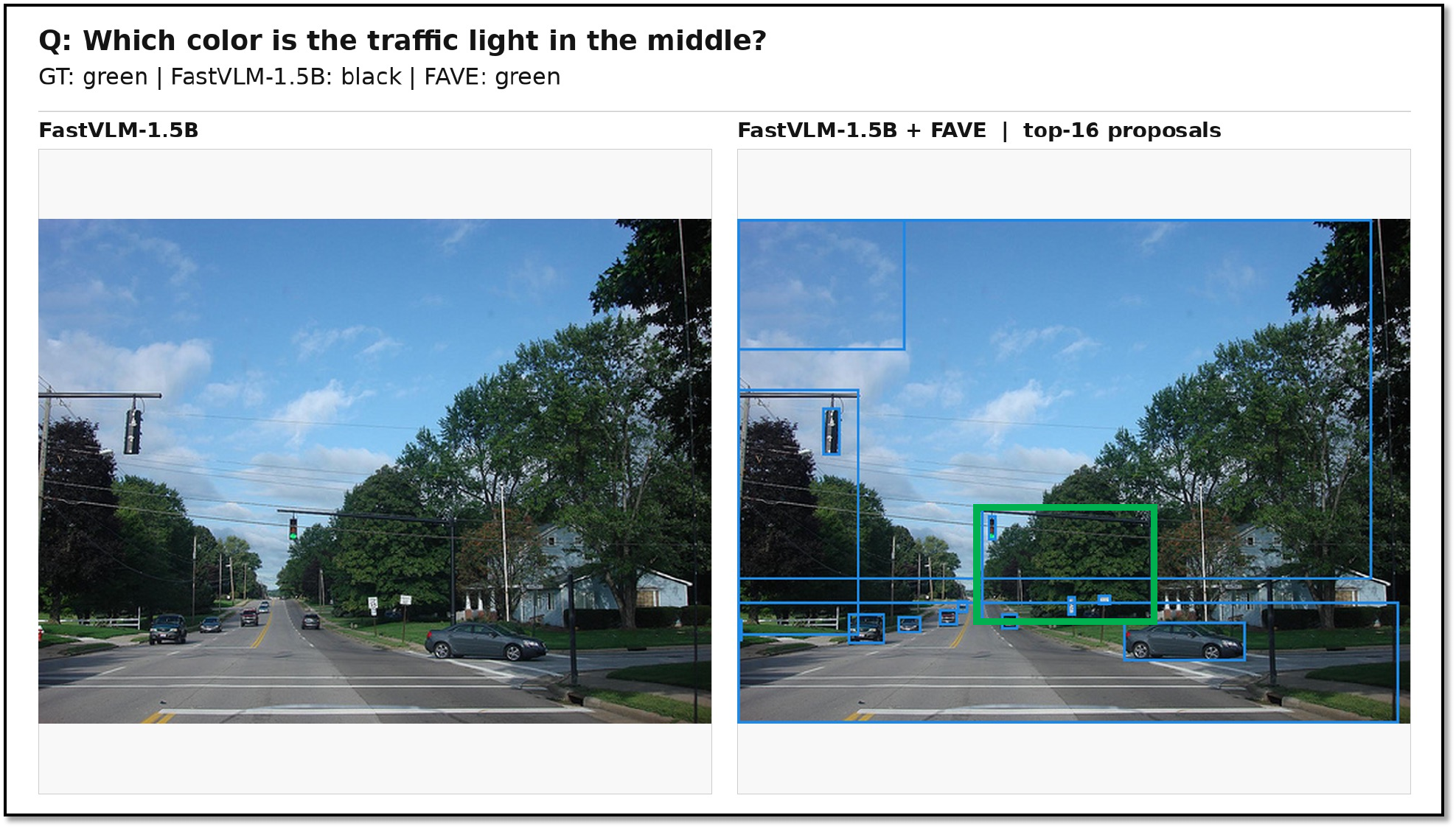}

    \caption{
    \textbf{Qualitative GQA examples of spatial disambiguation and small-object
    attribute recovery.}
    In the first example, the field is grassy but not white; the visible white
    regions belong to the players. FAVE correctly answers \emph{no}, separating
    attributes of nearby objects from those of the queried field. In the second,
    FAVE correctly identifies the middle traffic light as \emph{green}, recovering
    fine-grained evidence from a small object within a larger street scene.
    }
    \label{fig:gqa_qual_3}
\end{figure*}

\subsection{GQA Training and Evaluation Details}

The GQA experiment in the main paper is a targeted evaluation of
fine-grained object appearance rather than aggregate GQA reasoning. We use the balanced GQA splits, with 942,848 training questions and
132,062 validation questions. Final evaluation is performed on the
12,578-question testdev-balanced split, restricted to the 5,186 questions
labeled \emph{attribute} under the GQA semantic question type. These questions cover
properties such as color, material, size, state, and shape.

\paragraph{Stage-1 object and attribute supervision.}
Unlike the TextVQA local encoder, the GQA Stage-1 encoder is trained directly
from GQA scene-graph object crops to preserve general object and appearance
semantics. Each retained scene-graph object forms one training sample. Object
identity is supervised with masked cross-entropy, while attributes are
represented as grouped multi-label targets over seven semantic groups:
color, material, shape, size, texture/pattern, state, and part/clothing
detail. Attribute supervision is masked per group: a missing group is treated
as unknown rather than as an all-negative target. Within an observed group,
multiple positive attributes are permitted, and the active group losses are
averaged equally so that groups with larger vocabularies do not dominate the
objective.

We additionally use two augmented views of each object crop and enforce
same-object representation consistency. For each active reducer slot, the
corresponding representation from the other view is the positive example and
representations from other object crops serve as negatives. The complete
Stage-1 objective is
\begin{equation}
    \mathcal{L}_{\mathrm{S1}}
    =
    \mathcal{L}_{\mathrm{obj}}
    +
    \mathcal{L}_{\mathrm{attr}}
    +
    0.1\,\mathcal{L}_{\mathrm{con}},
\end{equation}
where $\mathcal{L}_{\mathrm{obj}}$ is masked object cross-entropy,
$\mathcal{L}_{\mathrm{attr}}$ is grouped masked binary cross-entropy, and
$\mathcal{L}_{\mathrm{con}}$ is a two-view InfoNCE loss with temperature
$0.07$. Stage-1 training samples the active reducer width uniformly from
$M\in\{1,2\}$; the downstream GQA system uses the nested $M=1$ prefix, so
each selected region is represented by one 192-D FAVE token.

\paragraph{Deferred long-tail balancing.}
GQA object identities and attributes are strongly long-tailed. We therefore
apply train-only logit adjustment to the two semantic losses while leaving the
raw model logits unchanged for validation and downstream use. For object class
$c$ with training frequency $n_c$, the training logit is
\begin{equation}
    \tilde{z}_c
    =
    z_c+\tau_{\mathrm{obj}}\log n_c .
\end{equation}
For attribute label $j$, let
\begin{equation}
    \pi_j
    =
    \frac{N_j^{+}}
         {N_j^{\mathrm{sup}}},
\end{equation}
where $N_j^{+}$ is its positive count and $N_j^{\mathrm{sup}}$ is the
number of training objects for which its attribute group is supervised.
The corresponding attribute logit is
\begin{equation}
    \tilde{z}_j
    =
    z_j
    +
    \tau_{\mathrm{attr}}
    \left[
        \log \pi_j-\log(1-\pi_j)
    \right].
\end{equation}
Balancing is deferred so that the representation first learns the unadjusted
semantic objective. During epochs 1--10,
$\tau_{\mathrm{obj}}=\tau_{\mathrm{attr}}=0$. From epochs 11--20 the
strengths increase linearly, reaching
$\tau_{\mathrm{obj}}=1.0$ and $\tau_{\mathrm{attr}}=0.5$ at epoch 20.
The Stage-2 model used in the reported experiment is initialized from this
epoch-20 Stage-1 encoder/reducer checkpoint.

Stage 1 is optimized with AdamW using a global batch size of 256, peak
learning rate $2.5\times10^{-4}$, weight decay $0.05$, four warm-up epochs,
and cosine learning-rate decay to $10^{-6}$. Crops follow the same
native-geometry FAVE preprocessing used downstream: they are not upsampled,
are aspect-ratio-preservingly downsampled only when the longest side exceeds
384 pixels, and are padded on the right and bottom to the 16-pixel patch grid.

\paragraph{Question-independent object proposals.}
Local regions are generated once per image using prompt-free YOLOE-26S and
are therefore independent of the GQA question. YOLOE operates at an input
size of 640 pixels with a confidence threshold of $0.001$ and retains at
most 1,000 candidate detections, initially ordered by descending detector
confidence. We then apply a conservative stable de-duplication policy
(\emph{Dedup-v2}). A proposal is considered a near-duplicate of an earlier
higher-ranked proposal only when their IoU is at least $0.85$ and their area
similarity,
\begin{equation}
    \frac{\min(A_i,A_j)}{\max(A_i,A_j)},
\end{equation}
is at least $0.70$. Non-duplicate proposals retain their original confidence
ordering. Near-duplicates are moved to the end rather than removed, preserving
the complete raw proposal pool. Dedup-v2 is deterministic and uses neither the
question text nor GQA labels, object-class semantics, or object-size quotas.

\paragraph{Stage-2 proposal composition and training.}
Stage 2 trains on the full balanced GQA training questions using the first
$K=32$ Dedup-v2 proposals for each image. The FastVLM-1.5B global vision
pathway and multimodal projector, Qwen2 language model and language head, and
the Stage-1 FAVE ViT are frozen. The $M=1$ Stage-1 reducer is fine-tuned and
a local projector mapping $192\!\rightarrow\!512\!\rightarrow\!1536$ is
trained to place the local FAVE tokens in the Qwen2 embedding space. The
resulting local tokens are concatenated after FastVLM's 256 global visual
tokens.

During training only, question-relevant GQA scene-graph objects are used to
guard against missing proposal coverage. Relevant objects are obtained from the union of object references in the
question annotation and semantic-program arguments. Resolved relevant GT objects are matched one-to-one to the selected
YOLOE proposals using deterministic maximum-cardinality matching with the
inclusive criterion $\mathrm{IoU}\geq0.50$. When a match exists, the YOLOE
proposal crop is retained; the GT crop does not replace it. Only an unmatched
resolved relevant object is inserted as a GT fallback. If the proposal set
already fills the $K=32$ budget, the fallback replaces the lowest-ranked
proposal that is not itself assigned to a relevant object. Thus the local
budget never exceeds 32 crops. GT boxes are used only for this training-time
fallback mechanism; validation and test-time inference use Dedup-v2 proposals
alone.

All proposal and fallback crops follow the same deterministic Stage-1
preprocessing contract: boxes are clamped to the image, converted to integer
crop bounds, extracted without upsampling, aspect-ratio-preservingly capped
at a longest side of 384 pixels, normalized with ImageNet statistics, and
right/bottom padded to the 16-pixel patch grid.

Stage 2 uses autoregressive answer supervision only. Prompt tokens are masked
from the loss and cross-entropy is applied to the assistant answer tokens.
No OCR, region-classification, or TextVQA-specific auxiliary objective is
used. The newly initialized local projector and the Stage-1 reducer are
optimized with AdamW at learning rates $10^{-4}$ and $10^{-5}$,
respectively, with a 3\% warm-up followed by cosine decay. Training runs for
two epochs, and validation teacher-forced answer cross-entropy is used for
checkpoint selection.

\paragraph{Inference and attribute evaluation.}
The reported operating point uses the same Stage-2 checkpoint but reduces
the inference proposal budget from $K=32$ to $K=16$. Specifically, we retain
the first 16 regions in the frozen Dedup-v2 ordering and keep $M=1$, yielding
at most 16 additional local visual tokens. This changes only the local
proposal/token budget: the FAVE encoder, reducer, local projector, FastVLM
global pathway, and language model weights are unchanged. No GT information
is available during inference.

Generation follows the matched FastVLM GQA protocol with the Qwen2
conversation template and the instruction
``Answer the question using a single word or phrase.'' We use greedy decoding
with one beam, no sampling, and at most 32 generated tokens. Predictions are
scored with the GQA evaluator, and the number reported in Table~7 of the main
paper is accuracy over the 5,186 testdev-balanced questions whose semantic
type is \emph{attribute}.

\paragraph{Targeted local capacity versus model scaling.}
The GQA comparison is intentionally restricted to this attribute subset. FAVE
increases the FastVLM-1.5B system from 1.675B to 1.682B parameters
($0.42\%$) while improving attribute accuracy from $70.73$ to $72.04$.
This matches the substantially larger Mini-Gemini-HD-8B ($72.02$) and
remains within $0.68$ points of FastVLM-7B ($72.72$). Consistent with the
main-paper interpretation, this result shows that targeted local visual
capacity provides a complementary route to fine-grained accuracy gains
without scaling the entire visual--language system.

\paragraph{Qualitative interpretation and connection to active vision.}
The qualitative examples make the role of the local pathway concrete. In each
case, answering correctly requires an attribute or state to be associated with
a particular object rather than inferred only from the global scene. The
traffic-light example is the clearest scale case: the queried light occupies a
small fraction of the street scene, and the global FastVLM pathway predicts the
wrong color, whereas the selected high-acuity crop preserves the local evidence
needed to identify it as \emph{green}. The plant-pot example exhibits the same
coarse-to-fine mechanism illustrated by the potted plant in Figure~1 of the
main paper. Global context establishes the object and surrounding scene, but
distinguishing whether the pot is \emph{full} or \emph{empty} depends on finer
state information within the object region; FAVE supplies this additional local
evidence.

The field example exposes a different failure mode: object--attribute binding.
The question asks whether the field is both \emph{grassy} and \emph{white}, so
both attributes must apply to the field itself. The global scene clearly
contains grass as well as salient white regions, but the selected local
proposals show that the white regions belong to the players. FAVE therefore
supplies object-local evidence that disambiguates where the white appearance
comes from, allowing the combined global--local representation to answer
\emph{no}. Here the global pathway provides the field-level scene context,
while the foveated pathway localizes the competing white evidence to the
players.

These cases also clarify the active-vision decomposition underlying FAVE.
The FastVLM global pathway provides passive scene encoding: it compresses the
complete image into a fixed global representation without allocating additional
acuity to particular regions. The local pathway separates spatial selection
from visual encoding. The question-independent proposal mechanism determines
\emph{where} additional processing can be allocated by identifying candidate
object regions, while FAVE determines \emph{what} visual evidence is preserved
from those regions through high-acuity adaptive encoding. The two pathways are
therefore complementary: the global branch preserves scene context, while the
foveated branch supplies additional object-centric evidence for fine appearance
and object--attribute binding.
{
    \small
    \bibliographystyle{ieeenat_fullname}
    \bibliography{main}
}

\end{document}